\documentclass[11pt]{article}

\usepackage[final]{acl}

\usepackage{times}
\usepackage{latexsym}
\usepackage{subfig}
\usepackage{tikz}
\usepackage{chngcntr}

\usepackage[T1]{fontenc}

\usepackage[utf8]{inputenc}

\usepackage{microtype}

\usepackage{inconsolata}
\usepackage{stfloats}

\usepackage{graphicx}
\usepackage{booktabs}

\usepackage[table]{xcolor}
\usepackage{mdframed}
\usepackage{multirow}
\usepackage{array}

\definecolor{werewolf}{RGB}{255, 204, 204}   
\definecolor{bargaining}{RGB}{229, 204, 255} 
\definecolor{coding}{RGB}{255, 229, 204}     
\definecolor{research}{RGB}{204, 229, 255}   

\title{Explicit Trait Inference for Multi-Agent Coordination}

\author{
  Suhaib Abdurahman$^1$\thanks{~Work done during internship at Amazon.}\quad Etsuko Ishii$^2$ \quad Katerina Margatina$^2$ \\[0.3em]
  {\bf Divya Bhargavi$^2$ \quad Monica Sunkara$^2\thanks{~Work done while at Amazon.}$ \quad Yi Zhang$^2$} \\[0.8em]
  $^1$University of Southern California, $^2$AWS Agentic AI Labs  \\[0.3em]
  {\texttt{sabdurah@usc.edu, \{etkishii, katemarg, yizhngn\}@amazon.com}}
}

\begin{document}
\maketitle
\begin{abstract}
    LLM-based multi-agent systems (MAS) show promise on complex tasks but remain prone to coordination failures such as goal drift, error cascades, and misaligned behaviors. 
    We propose \textbf{Explicit Trait Inference (ETI)}, a psychologically grounded method for improving coordination. 
    ETI enables agents to infer and track partner characteristics along two established psychological dimensions—\emph{warmth} (e.g., trust) and \emph{competence} (e.g., skill)—from interaction histories to guide decisions. 
    We evaluate ETI in controlled settings (economic games)
    , where it reduces payoff loss by 45–77\%, and in more realistic, complex multi-agent settings (MultiAgentBench), where it improves performance by 3–29\% depending on the scenario and model, relative to a CoT baseline. 
    Additional analysis shows that gains are closely linked to trait inference: ETI profiles predict agents’ actions, and informative profiles drive improvements. 
    These results highlight ETI as a lightweight and robust mechanism for improving coordination in diverse multi-agent settings, and provide the first systematic evidence that LLM agents can (i) reliably infer others’ traits from interaction histories and (ii) leverage structured awareness of others’ traits for coordination.
\end{abstract}

\section{Introduction}
Large language model (LLM) based multi-agent systems (MAS) promise to tackle tasks too complex for single agents by distributing work across specialized collaborators \citep{guo2024large, talebirad2023multi, tran2025multi} and enable applications that inherently require multiple actors, such as world simulations \citep{qian2024scaling, huang2023agentcoder, park2023generative, gao2024large, xiao2023simulating}. Despite this potential, MAS remain prone to failures of coordination, such as goal drift, error cascades, insufficient information sharing, and misaligned actions, that undermine reliability and limit scalability \citep{han2024llm, cemri2025multi}. Thus, a central challenge is not whether agents can execute individual actions, but whether they can coordinate effectively with each other.

Human psychology offers a roadmap for improving agent coordination. Decades of research show that \emph{social evaluations}, the formation of impressions of others, strongly influence how humans coordinate \citep{fiske2007universal}. Grounded in task-relevant observations—such as contribution quality, willingness to help, or information sharing—social evaluations improve coordination and performance \citep{hackel2020reinforcement, derfler2022swiftly} by helping humans determine whom to trust, whom to allocate tasks to, and anticipate others’ behavior \citep{mayer1995integrative, faraj2000coordinating,  hinds2000choosing, cuddy2011dynamics, balliet2013trust, marjieh2024task}. 

Psychology research further provides well-established taxonomies for organizing social evaluations, most prominently the two dimensions of \emph{warmth} (e.g., cooperativeness, integrity) and \emph{competence} (e.g., skill, reliability) \citep{fiske2007universal, abele2014communal}. These dimensions offer a compact and widely validated basis for representing others in coordination-relevant terms, making them especially useful for modeling partners in multi-agent contexts.

A particularly promising form of social evaluation for MAS is \emph{trait inference}—the attribution of relatively stable behavioral dispositions to others \citep{jones1965acts}. In multi-agent contexts, such inferences can be grounded in behavioral evidence from agent instructions, planning, communication, and interaction history, rather than superficial cues like appearance or demographics that in humans are often linked to bias and stereotypes \citep{fiske2015intergroup, cuddy2008warmth}. This grounding makes trait inference both task-relevant and less prone to bias, providing a natural way to model others’ in support of coordination (e.g., recognizing that a partner is capable but uncooperative may guide task allocation or feedback).

\begin{figure*}[!t]
  \centering
  \includegraphics[width=1\textwidth]{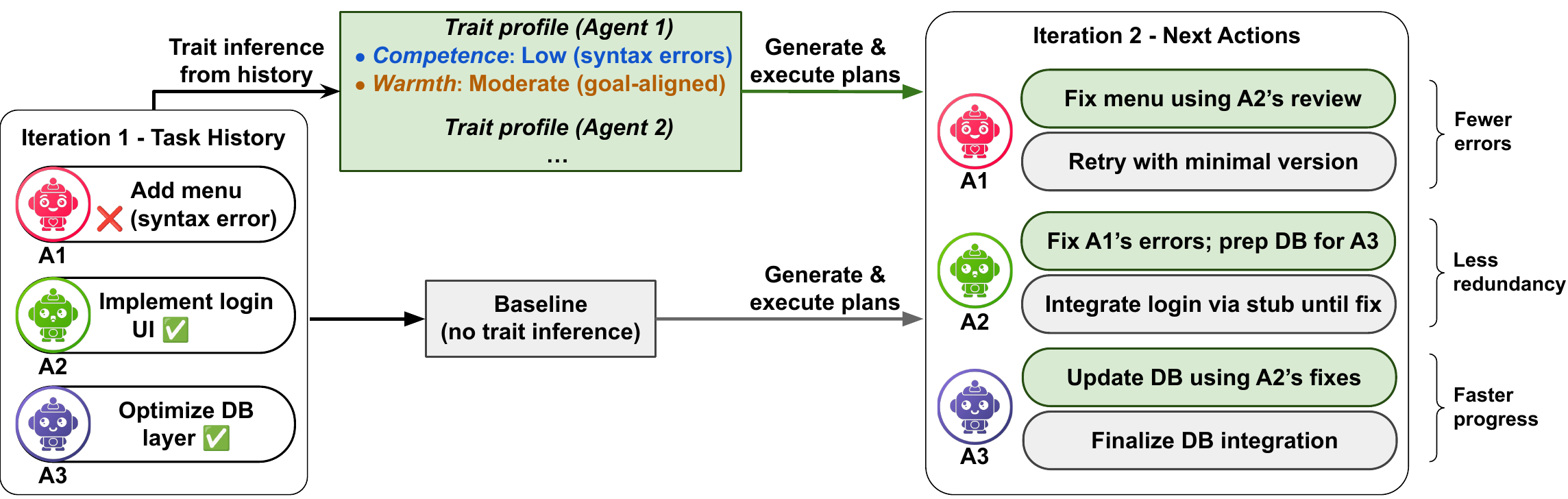}
  \caption{\textit{ETI improves agent coordination}.
\textbf{Left:} three agents collaborate to develop a web-app. Each handles a subtask; one fails.
\textbf{Middle:} ETI infers competence and warmth from history, while baseline skips this step.
\textbf{Right:} Trait-aware agents plan around others’ strengths, weaknesses, and goals, while baseline agents follow generic coordination focused on the immediate failure. By grounding next steps in inferred traits (e.g., A2 is skilled, A1 is unreliable), ETI steers agents to build on high-quality work, cover weak spots, and avoid duplicating effort.}
  \label{fig:overview}
\end{figure*}

Building on these insights, we propose \textbf{Explicit Trait Inference (ETI)}, a framework that scaffolds agents with mechanisms for generating and maintaining representations of others’ traits. Agents generate structured profiles in terms of warmth and competence, update them as interactions unfold, and use them to guide decisions (Figure~\ref{fig:overview}).

ETI targets MAS failures by linking trait inference to concrete coordination adjustments:
\begin{itemize}
\item \textbf{Warmth} addresses \emph{goal drift, unreliable cooperation, or adversarial behavior}. Low warmth prompts agents to clarify intentions, supply missing information, discount unreliable input, or guard against suspicious partners.
\item \textbf{Competence} addresses \emph{execution errors, cascading failures, and poor adaptability}. Low competence prompts agents to reassign tasks,  adjust tasks, or increase verification.
\end{itemize}
In this way, interaction histories are distilled into stable trait profiles that surface coordination risks and opportunities at decision time, allowing agents to systematically adapt planning, delegation, and communication.

We evaluate ETI in two stages: first, using economic games as controlled, interpretable testbeds with ground truth traits to assess inference accuracy, robustness, and effects on decision-making; second, on MultiAgentBench \citep{zhu2025multiagentbench} to test generalization to more complex, realistic MAS.

ETI proves effective across both settings. In economic games, ETI agents make more optimal decisions, reducing payoff (game rewards) loss by 42–77\%. On MultiAgentBench, spanning collaborative and competitive scenarios, ETI improves performance by 3–29\% and coordination by 6–42\%, depending on the model and scenario. Supplemental analyses show that these gains stem from ETI itself: trait inferences predict agents’ actions, and only diverse, informative profiles yield improvements, whereas generic ones provide little benefit.

Our results provide the first systematic evidence that (1) LLM agents can reliably infer others’ traits from interaction histories, and (2) ETI improves coordination and performance in MAS by giving agents structured awareness of their partners’ traits. 
By grounding multi-agent reasoning in well-established dimensions of social cognition, ETI opens a path toward MAS that are not only more effective but also more transparent and theoretically motivated.

\section{Related Work}

\subsection{Social Traits in LLMs}
A growing line of research examines the psychology of LLMs to understand their behavior in social and collaborative settings. Several studies find that LLMs exhibit stable behavioral patterns and personality structure \citep{bhandari2025evaluating, frisch2024llm, li2025llm, leng2023llm, chen2025persona, jiang-etal-2024-personallm}. In human–AI interactions, humans can detect and evaluate these patterns along psychological dimensions of warmth and competence \citep{mckee2024warmth, xie2024can}. Moreover, LLM agents themselves can adopt and maintain distinct personas, displaying behaviors such as cooperativeness, agreeableness, or sycophancy \citep{park2024generative, samuel2024personagym, sun2024building, fanous2025syceval}.
ETI builds on these findings to provide agents with representations of others’ stable characteristics that can inform coordination. However, it does not rely on LLMs having intrinsic human-like traits. Instead, it leverages any stable, task-relevant behavioral patterns—such as repeated failures or goal drift—by encoding them as traits to guide coordination decisions.

\subsection{Improving Coordination in MAS}
\paragraph{Structural Approaches.} Structural approaches improve coordination in MAS by organizing workflows and agent structures to reduce errors and enhance performance. Mixture-of-agents and ensemble methods combine diverse models to exploit complementary strengths \citep{wang2024mixture, guo2024large}, while frameworks like CAMEL, ChatDev, AutoGen, and hierarchical ``challenger/inspector’’ schemes structure collaboration through explicit roles (e.g., coders, reviewers, monitors) \citep{li2023camel, qian2023chatdev, wu2024autogen, huang2024resilience}. However, these approaches do not address how agents reason about or adapt to one another—for example, assessing trustworthiness or deciding how to respond. ETI instead provides agents with coordination-relevant information about partners, enabling flexible, adaptive interaction without imposing fixed workflows.

\paragraph{Structured Reasoning.} Various approaches improve LLM performance by scaffolding their reasoning processes. Chain-of-Thought \citep[CoT;][]{wei2022chain} prompting guides models to externalize intermediate reasoning steps, making it more accurate and interpretable \citep{wei2022chain, yao2023tree, wang2023plan}. Reflective methods such as ReAct \citep{yao2023react} and Reflexion \citep{shinn2023reflexion} similarly structure agents’ reasoning to reduce errors and improve downstream coordination \citep[e.g.][]{bo2024reflective, song2025gradientsys}. ETI extends this idea to the \emph{social domain}, providing a structured scaffold for reasoning about other agents via trait profiles rather than purely task-level deliberation.

\paragraph{Social Reasoning.} 
Relatedly, a growing body of work explores how agents can reason about others to improve coordination. Theory-of-Mind approaches prompt agents to model others’ knowledge, beliefs, or intentions \citep{li2023theory, guo2023suspicion, richards2024you, sclar2023minding}, while partner and opponent models forecast others’ behavior to support negotiation or deception detection \citep{yu2025llm, meta2022human}. However, these methods primarily model transient mental states (e.g., beliefs) rather than stable traits like reliability or cooperativeness. While trait patterns may sometimes be implicitly reflected in those models, they do not explicitly represent or track them. ETI complements these approaches by directly modeling such traits across interactions. 

\paragraph{Reputation Systems.} Reputation systems and adjacent methods, such as credit assignment, improve coordination by tracking agents’ past behavior. Reputation mechanisms estimate factors such as success rates to guide partner selection \citep{lou2025drf}, while credit-assignment methods attribute contributions to team outcomes to aid multi-agent learning \cite{lin2025speaking}. However, these approaches typically focus on task-specific outcome metrics (e.g., success rates, rewards), capturing \emph{what} agents achieve but not \emph{why} or \emph{how}. ETI, by contrast, builds psychology-grounded profiles that integrate goals, intentions, actions, and outcomes, providing richer representations for deciding both \emph{whom} to collaborate with and \emph{how} to coordinate

\paragraph{LLMs in Game-Theoretic Settings.}
Within the broader game-theoretic literature, our work aligns with research that uses classic economic games to probe LLMs’ social behavior and cooperativeness. Recent benchmarks and surveys study how LLMs behave in settings such as Prisoner’s Dilemma, documenting stable but often suboptimal social strategies, biases in cooperation and reciprocity, and difficulties when coordination requires active reasoning about partners \citep[e.g.,][]{sun2025game, akata2025playing, herr2024large, wang2024tmgbench, agashe-etal-2025-llm}. Related work on communication-based games likewise finds human-like patterns of trust, cooperation, and deception in LLM agents \citep{xu2023exploring, wang2023avalon}. ETI builds on these insights by providing an explicit, behavior-based partner model from interaction histories, with our economic-game environments serving as controlled testbeds to evaluate the accuracy and utility of these trait inferences for decision-making.

\section{Methodology}
\subsection{Trait Framework}
Our framework represents agents along the warmth and competence dimensions from social psychology. However, to avoid conflating these abstract dimensions with colloquial understandings, we operationalize each dimension through four concrete, behaviorally anchored traits (Table~\ref{tab:traits}), adapted from prior work on trait terms and social cognition \citep[e.g.,][]{abele2008towards}. Traits were chosen to map strongly onto their respective dimensions, remain conceptually distinct, and were defined in clear behavioral terms for use in prompting. A key design choice was to make the warmth–competence separation explicit in the definitions, countering the colloquial tendency to conflate traits (e.g., judging someone uncooperative as incompetent). For example, \emph{Ability} is defined solely in terms of successful execution of intended actions, independent of intentions. This prevents traits from collapsing into a single undifferentiated signal and yields clearer, more informative representations aligned with the psychological framework.

\begin{table}[t]
\centering
\small
\begin{tabular}{p{1.25cm}p{5.75cm}}
\toprule
\textbf{Dimension} & \textbf{Traits (Short Description)} \\
\midrule
Warmth & \emph{Goal Alignment}: work towards task goals \\
       & \emph{Collaboration}:  coordinate; share relevant info \\
       & \emph{Trustworthiness}:  avoid deception; be honest. \\
       & \emph{Maliciousness}:  undermine/sabotage goals \\
\midrule
Competence~ & \emph{Ability}:  execute actions successfully \\
           & \emph{Reliability}:  deliver consistent quality \\
           & \emph{Adaptability}:  adjust to changing contexts \\
           & \emph{Efficiency}:  minimize resources/time \\
\bottomrule
\end{tabular}%
\caption{Eight behaviorally anchored traits instantiating warmth and competence. Full definitions and prompt templates are provided in the Appendix (\ref{app:trait_definitions_prompts}).}
\label{tab:traits}
\end{table}

\subsection{Explicit Trait Inference Procedure}
\begin{figure*}[!t]
  \centering
  \includegraphics[width=1\textwidth]{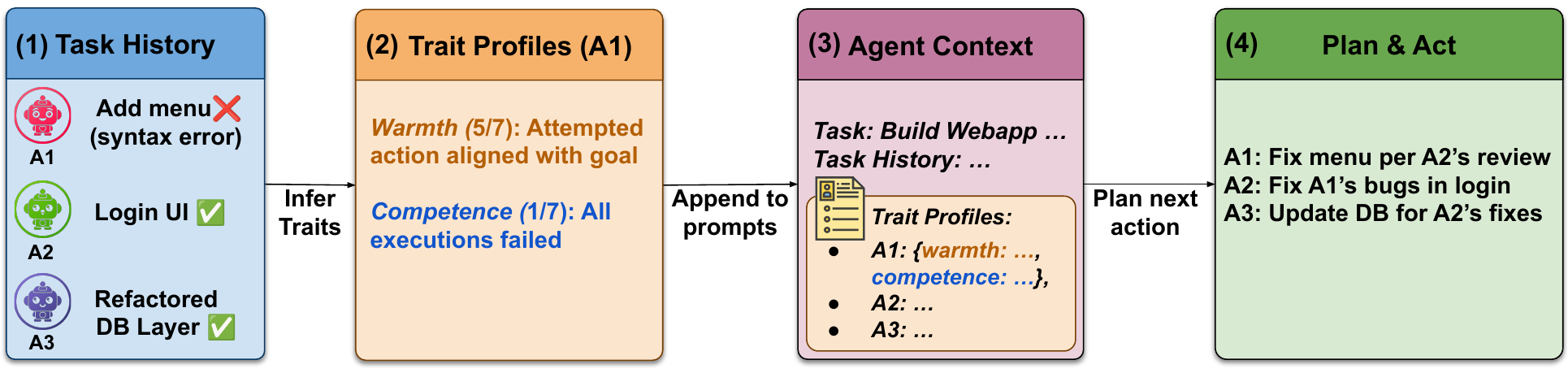}
  \caption{Flow of \textit{Explicit Trait Inference} within a multi-agent task.
After each iteration, agents (1) draw on the history of goals, planned subtasks, and observed results 
to (2) infer structured trait profiles from partners’ behavior, 
(3) incorporate profiles into their \emph{agent context}, 
and (4) use the enriched context to plan and execute next steps.}
  \label{fig:trait_flow}
\end{figure*}

We implement ETI as a prompting and context-management procedure. After each interaction, agents receive structured summaries of task goals, actions, communication, and outcomes, and are prompted to infer partners’ traits. Agents produce (a) 1–7 Likert ratings for each trait—a standard format for social evaluations in psychology \citep[e.g., see ratings in][]{fiske2007universal, abele2008towards}—and (b) concise evidence supporting each judgment, and these profiles are appended to their context for subsequent planning and execution. Prompts instruct models to focus on dominant behavioral patterns rather than isolated events and avoid task-specific decision rules, which stabilizes inferences while keeping the setup minimal and domain general. This prompting-based design requires no fine-tuning or additional data and adds only minimal overhead, making ETI lightweight and broadly applicable. Figure~\ref{fig:trait_flow} illustrates this integration within a generic multi-agent pipeline.

\paragraph{Illustrative schema.} The instructions for trait inference followed this schema:

\begin{quote}\small
\textbf{Scenario:} \emph{[description of task goal and rules]}\\[0.3em]
\textbf{Task context:} \emph{[own actions, partners’ actions, outcomes, additional info]}\\[0.5em]
\textbf{Instruction:} Based on the above, determine the extent to which the following traits apply to \{PARTNER\_NAME\}.\\[0.5em]
\textbf{Output format:}\\
\verb|{|
\verb|  "trait": rating (1--7),|
\verb|  "evidence": "..."|
\verb|}|
\end{quote}

Full templates and examples are provided in the Appendix (\ref{app:trait_definitions_prompts}).

\section{Economic Games}
\label{sec:methods_econ}

Economic games offer a controlled, interpretable testbed to assess whether ETI provides accurate trait inference and improves decision-making. Their simple rules and reward (``payoffs'') mechanisms create decision problems that are easy to analyze yet require adaptive reasoning to perform well. Controlling the game (e.g., opponents’ traits) lets us create ground truth to test agents’ trait inference and decision-making.

\subsection{Set-Up}
We evaluate ETI on two games: the \textbf{Iterated Prisoner’s Dilemma} (PD) and the \textbf{Iterated Stag Hunt} (SH). 
Both are binary-choice tasks where the combination of agents’ decisions determines their rewards according to a payoff matrix (see Appendix \ref{app:econ_game_section}). PD mixes cooperation and competition, requiring agents to weigh mutual benefit against exploitation, while SH is purely collaborative, trading a safe individual payoff for a higher one that succeeds only if both players coordinate. Together, these games create settings where optimal actions depend on opponent traits and context, requiring adaptive strategies rather than simple heuristics.

\paragraph{Competence-based payoffs.} Standard PD and SH games assume that players' choices are always successfully executed. We add a \emph{competence} parameter: a player realizes the intended payoff only with probability $p_i$. 
This lets agents infer (a) \emph{intent} from the action (cooperative vs.\ selfish) and (b) \emph{ability} from whether the outcome matches the expected payoff. 
Formally, if player $i \in \{1,2\}$ chooses $a_i \in \mathcal{A}$, their action succeeds with $s_i \sim \mathrm{Bernoulli}(p_i)$. In case of failure, the base payoff $M(a_i,a_j)$ is adjusted to $M(a_i,a_j,s_i,s_j)$, accounting for any agents' execution failures.

\begin{figure}[t!]
  \centering
  \includegraphics[width=1\columnwidth]{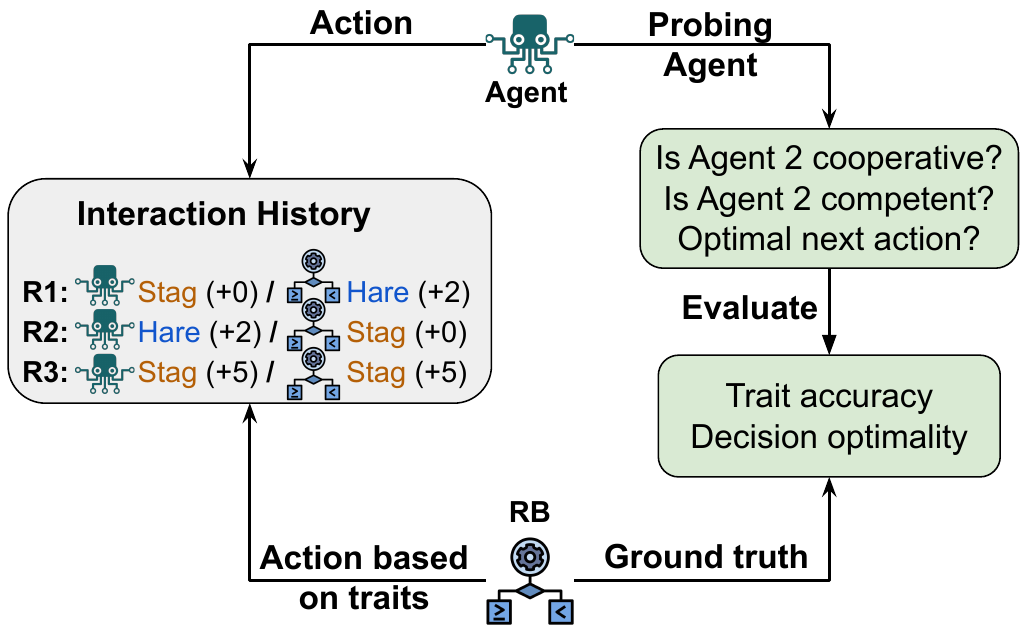}
\caption{Evaluation loop (Stag Hunt example). An agent plays against a scripted opponent (RB) with fixed traits. Each round, the agent observes the interaction history and realized payoffs, then selects its next action (ETI agents first generate explicit trait profiles, baseline agents do not). In parallel, we probe the agent about the opponent’s traits. We evaluate (i) probe accuracy against ground truth and (ii) decisions against the payoff-optimal response.}
\label{fig:econ_games}
\end{figure}

\paragraph{Rule-based opponents.} To ensure interpretable ground truth, each run pairs an LLM agent with a scripted opponent. The opponent is parameterized by two probabilities: (a) \emph{warmth}, the chance of choosing the cooperative action ($p_x$), and (b) \emph{competence}, the chance of executing it successfully ($p_i$). 
Actions are drawn from $p_x$ and outcomes are drawn from $p_i$. 
Competence applies only to the scripted opponent; LLM agents always successfully execute ($s_1=1$). Figure~\ref{fig:econ_games} illustrates the interaction and evaluation loop.

\subsection{Experiments}
We use \textsc{Qwen}3-8B, a lightweight reasoning model, as the acting agent in all runs. 

\paragraph{Design.} Each run pairs agents with a scripted opponent for 50 rounds, allowing behavioral evidence to accumulate. We vary opponent \emph{competence} ($p_i \in {0,1}$) and \emph{warmth} ($p_x \in {0,1}$), and additionally test a \emph{noisy} variant with ($p_i, p_x \in {0.15,0.85}$) to simulate settings with inconsistent behavior and unclear signals. We run every configuration 25 times. This setup lets us evaluate whether ETI (i) provides accurate trait inference, (ii) improves decision quality, and (iii) remains robust to noise.

\paragraph{Evaluation.} For \textit{trait inference accuracy}, agents are probed each round—using the same interaction history and prompt as for action generation—to predict whether the opponent is cooperative and competent. Predictions are compared with the opponent’s ground-truth (e.g., ``cooperative’’ if $p_x > 0.5$, ``competent’’ if $p_i > 0.5$) and scored with F1. For \textit{decision quality}, we treat each round as a binary classification task and compute the F1 of the agent’s action against the payoff-optimal response (defined from the opponent’s parameters and the payoff matrix). Additionally, we measure the percentage deviation of the agent’s cumulative payoff from that of the optimal strategy (lower is better). All metrics are reported as means across runs. Differences between ETI and baseline are tested for statistical significance using independent-samples $t$-tests ($p < 0.05$; significance indicated by asterisk (*)).

\paragraph{Baseline.} We use a CoT-style baseline in which the agent receives the game description and history of past decisions and outcomes, and is instructed to reason about this information to select its next action. We thus test whether ETI leads to improvements \textit{above and beyond} general reasoning about others' past actions.

\subsection{Results} \label{res:econ_games}
\paragraph{Trait inference accuracy.} ETI provides more accurate trait inference than implicit modeling. In the Prisoner’s Dilemma (PD), competence F1 rises from 0.69 to 0.89* and cooperation F1 from 0.43 to 0.73*. In Stag Hunt (SH), competence F1 is already high (0.85 to 0.86), but cooperation F1 improves substantially from 0.52 to 0.81*. See a full overview in Table \ref{tab:trait-inference}.

\begin{table}[t]
\centering
\setlength{\tabcolsep}{3pt}       
\renewcommand{\arraystretch}{0.9} 
\begin{tabular}{@{}ll lcc@{}} 
\toprule
Game & Condition & Trait & Baseline & \multicolumn{1}{c}{\shortstack{ETI}} \\
\midrule
PD & Overall & Comp. & 0.69 & \textbf{0.89*} \\
    &         & Coop. & 0.43 & \textbf{0.73*} \\
    & Noisy   & Comp. & 0.06 & 0.71* \\
    &         & Coop. & 0.02 & 0.49* \\
\addlinespace
SH  & Overall & Comp. & 0.85 & 0.86* \\
    &         & Coop. & 0.52 & \textbf{0.81*} \\
    & Noisy   & Comp. & 0.88 & \textbf{0.90} \\
    &         & Coop. & 0.34 & 0.80* \\
\bottomrule
\end{tabular}
\caption{Trait inference accuracy (F1) averaged across all (Overall) and noisy opponent conditions (Noisy). Results are reported for competence (Comp) and cooperation (Coop). Bold indicates the best performance per trait and game; asterisks (*) mark statistically significant improvements over baseline ($p < 0.05$).}
\label{tab:trait-inference}
\end{table}

\paragraph{Robustness to noise.} ETI is substantially more robust to noisy signals than implicit modeling. In SH, ETI maintains high accuracy (F1 = 0.90 competence; 0.80* cooperation), while the baseline drops for cooperation (0.34). In PD, ETI remains moderate to high (0.71* competence; 0.49* cooperation) as the baseline collapses (0.06; 0.02).

\paragraph{Decision optimality.} ETI yields substantially higher optimal choice accuracy. In PD it improves F1 from 0.71 to 0.82*, and in SH from 0.20 to 0.75*. These gains indicate that agents equipped with ETI not only infer their partners’ traits more accurately, but also adjust their strategies accordingly, producing more optimal decisions.

\begin{figure}[t]
  \includegraphics[width=.9\columnwidth]{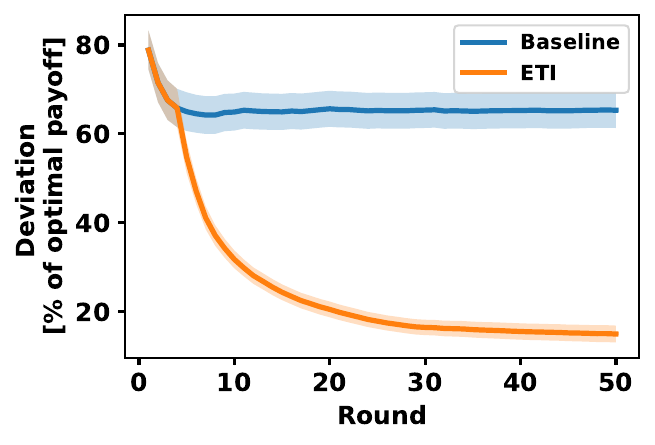}
  \caption{Relative deviation from optimal payoff across rounds (lower is better). ETI yields less deviation than the baseline after only $\approx$5 rounds. Shaded area reflects 95\% confidence interval.}
  \label{fig:deviation_staghunt}
\end{figure}

\paragraph{Payoff optimality.} In both games, ETI significantly reduces deviation from the optimal payoff. As shown in Figure \ref{fig:deviation_staghunt}, ETI agents rapidly approach the optimal strategy in SH after only a few rounds of evidence collection, while the baseline continues to lag (65\% vs.\ 15\%* deviation; a 77\% reduction in payoff-loss). A similar pattern holds in PD (135\% vs.\ 78\%* deviation; see Appendix Fig.~\ref{fig:deviation_prisoner}).

\medskip
Taken together, the results show that (i) LLM agents can reliably infer traits from interactions, (ii) ETI substantially improves robustness under noise, and (iii) these gains translate into more payoff-optimal decisions. Supplementary analyses (Appendix \ref{appendix: warmup}, \ref{appendix: generalizability}, \ref{app:adaptability}) further show that, after initial evidence collection, ETI can be applied without continuous updates, enabling more efficient use; that trait profiles transfer across tasks, supporting generalizable partner modeling; and that, in some cases, ETI helps agents adapt to sudden changes in partner characteristics. 

\section{Multi-Agent Benchmark}
Multi-agent benchmarks let us test whether ETI generalizes from controlled games to complex, realistic settings spanning collaborative, competitive, and adversarial dynamics. They involve more agents, richer interactions, and broader action spaces, providing a stronger and more comprehensive test of ETI’s performance across diverse interaction types.

\subsection{Set-Up}
\label{sec:methods_bench}
We evaluate ETI on the \textbf{MultiAgentBench} benchmark\footnote{The MultiAgentBench framework is publicly available and distributed under the MIT License.}, which spans diverse multi-agent tasks, provides structured action spaces (e.g., code editing, research tools, role-based communication), and includes predefined metrics for task success and coordination quality. We focus on four representative scenarios:

\paragraph{Coding.} Three agents collaborate on a shared codebase, either peer-to-peer or under a central planner, using file inspection, editing, reviewing, and messaging to coordinate.
\paragraph{Research.} Five agents co-author a research proposal, querying external APIs, inspecting PDFs, editing text, and discussing ideas.
\paragraph{Bargaining.} Buyers and sellers negotiate over a product in a non-zero-sum setting, exchanging offers, counters, and terms, accepting, walking away, or switching partners.
\paragraph{Werewolf.} Nine agents play a social-deduction game with open discussion and role-specific actions (e.g., accusation, protection, elimination) under asymmetric information and deception.

\subsection{Experiments}

\paragraph{Design.} We run each scenario 100 times with random seeds. We test both \textsc{Qwen}3-8B and \textsc{GPT}-4o-mini (as in~\citet{zhu2025multiagentbench}; default inference parameters) as agent models. In addition to within-model conditions, we include cross-model conditions where one model generates actions and the other executes ETI. Our main analysis compares the performance and coordination of ETI and baseline agents as well as the effects of mixing agent and trait inference model.

\paragraph{Evaluation Metrics.} Evaluation follows the benchmark scoring: (i) \emph{Task Performance}, combining task success, progress on milestones, and quality assessments appropriate to each domain, and (ii) \emph{Coordination Quality}, which combines inter-agent communication, planning, and reasoning regarding effective coordination. We report these metrics as defined in the benchmark framework. Improvements of ETI over baseline are tested for statistical significance using independent-samples $t$-tests ($p<0.05$; indicated with asterisks (*)). 

\paragraph{Baseline.} As a baseline, we adopt the benchmark’s task-specific CoT scaffolding, which guides agents through domain-relevant reasoning steps. For example, in the \emph{Coding} task, agents are instructed to first identify issues, then devise strategies, and finally implement fixes. In the \emph{Werewolf} game, agents are prompted to consider their role, analyze past actions of others (e.g., to detect contradictions), and generate intermediate outcomes such as suspect lists or next-step plans. These task-specific CoT instructions ensure that agents already reason about the problem in a systematic, domain-relevant way. Our experiments therefore test whether ETI improves coordination and performance \emph{above and beyond} task-relevant reasoning.

\begin{table*}[t]
\centering
\small
\begin{tabular}{ll
>{\columncolor{werewolf}}c >{\columncolor{werewolf}}c
>{\columncolor{bargaining}}c >{\columncolor{bargaining}}c
>{\columncolor{coding}}c >{\columncolor{coding}}c
>{\columncolor{coding}}c >{\columncolor{coding}}c
>{\columncolor{research}}c >{\columncolor{research}}c}
\toprule
 & & \multicolumn{2}{c}{Werewolf} & \multicolumn{2}{c}{Bargaining} & \multicolumn{2}{c}{Coding (Graph)} & \multicolumn{2}{c}{Coding (Tree)} & \multicolumn{2}{c}{Research} \\
\cmidrule(lr){3-4} \cmidrule(lr){5-6} \cmidrule(lr){7-8} \cmidrule(lr){9-10} \cmidrule(lr){11-12}
Agent & Trait Source & Task & Coord. & Task & Coord. & Task & Coord. & Task & Coord. & Task & Coord. \\
\midrule
\rowcolor{lightgray} \textsc{GPT}  & none & 25.26 & 54.32 & 57.16 & 77.48 & 51.21 & 57.01 & 45.79 & 52.22 & 69.83 & 56.36 \\
\textsc{GPT}  & \textsc{GPT}  & 29.54 & 55.56 & 57.04 & 70.12 & 53.31* & 73.52* & 50.69* & 72.95* & 68.32 & 73.18* \\
\textsc{GPT}  & \textsc{Qwen} & 36.46* & 57.56* & 58.83 & 69.65 & 52.84* & 74.38* & 50.22* & \textbf{74.43*} & 70.31 & 69.76* \\
\addlinespace
\rowcolor{lightgray} \textsc{Qwen} & none & 43.28 & 60.20 & 59.11 & 74.98 & \textbf{57.41} & 74.29 & 45.75 & 60.41 & 62.84 & 70.63 \\
\textsc{Qwen} & \textsc{GPT}  & 49.97 & 59.52 & 55.74 & \textbf{88.78*} & 56.44 & 84.43* & 52.13* & 66.81* & \textbf{75.99*} & 77.06* \\
\textsc{Qwen} & \textsc{Qwen} & \textbf{55.75*} & \textbf{65.20*} & \textbf{60.48*} & 79.43 & 56.82 & \textbf{86.46*} & \textbf{52.98*} & 66.34* & 69.98* & \textbf{78.35*} \\
\bottomrule
\end{tabular}
\caption{Results of the four scenarios averaged over 100 runs, shaded by color. Baselines (\texttt{none}) are greyed out. ETI improves coordination (Coord) and performance (Task), with the strongest gains from \textsc{Qwen}-generated profiles, especially for \textsc{Qwen} agents. \textit{(Asterisks (*) indicates statistically significant improvement over baseline, $p < 0.05$)}}
\label{tab:overall_marble}
\end{table*}

\subsection{Main Results}
Across all four scenarios, ETI improves both coordination and task performance (Table~\ref{tab:overall_marble}).  
Gains appear for \textsc{Qwen} and \textsc{GPT} agents alike, with the largest effects when trait profiles are produced by \textsc{Qwen}, independent of the agent model. Cross–model runs (e.g., \textsc{GPT} agents using \textsc{Qwen}-generated profiles) suggest that these improvements stem from the \emph{specific content of the profiles}, not merely from adding any structured note or instruction: a \textsc{GPT} agent given profiles by \textsc{Qwen} achieves large gains (e.g., Coding–Graph coordination $57.01\!\to\!74.38$*, Werewolf task $25.26\!\to\!36.46$*), whereas \textsc{GPT}-generated profiles often provide little or no benefit—and can even reduce performance (e.g., Bargaining task for \textsc{Qwen} agents $59.11\!\to\!55.74$). Gains in task performance vary with scenario and are strongest where the environment supports adaptive task allocation, partner selection, or adversarial reasoning. For example, Coding–Tree (\textsc{GPT} $45.79\!\to\!50.69$*, \textsc{Qwen} $45.75\!\to\!52.98$*), Research (\textsc{Qwen} $62.84\!\to\!75.99$*), and Werewolf (\textsc{GPT} $25.26\!\to\!36.46$*). Coordination improves broadly across all settings. This shows that ETI can effectively improve coordination and performance across diverse scenarios, but effect sizes vary by scenario and model.

\subsection{Additional Metrics}
Sub-scores for planning, communication, and milestone completion mirror the main trends. For instance, ETI agents produce more concrete plans (e.g., Coding–Graph, planning +9.1*), exchange more relevant messages (communication +15.2*), and reach \emph{shared} milestones more often (e.g., in Bargaining, contributors per milestone +8\%*). In settings where dialogue and incremental progress are key (e.g., negotiation, research), these sub-scores rise in tandem with coordination, indicating that ETI translates to practical, task-oriented behavior. Additional metrics (e.g., planning scores, KPIs) are reported in Appendix~\ref{app:add_metrics}.

\subsection{Trait–driven Mechanisms}
\paragraph{ETI Guides Agent Decisions.} In \textsc{Werewolf}, where faction labels provide ground truth, we conducted targeted analyses to test whether ETI directly shapes agent behavior (see details in Appendix~\ref{app:trait_dist_marble}). Trait profiles clearly separate villagers from werewolves: villagers score higher on trustworthiness ($\Delta=+0.75$ on the 1–7 scale)) and lower on maliciousness ($\Delta=-0.73$). These separations are sharper for \textsc{Qwen} than for \textsc{GPT} ($\Delta_\mathrm{model}=+0.15$, $-0.27$), aligning with \textsc{Qwen}’s superior performance. Crucially, the gaps widen in winning games ($\Delta_\mathrm{win}=+0.87$ trust, $-0.71$ maliciousness), showing that agents exploit clearer signals to detect adversaries and that stronger separation predicts success. Finally, to test temporal causality, we fit logistic models predicting key actions from prior trait inference: higher maliciousness ratings increases the chance of an agent being voted out (i.e., flagged as an adversary) ($\beta=+2.74$*), while higher trustworthiness reduces it ($\beta=-5.25$*). Probing transcripts confirms that agents incorporate trait inference when deciding (e.g., \textit{``Ethel’s persistent focus on Mae and high maliciousness rating suggest she may be trying to deflect suspicion; I should target her.''}; \ref{fig:eti_reasoning_example}). These patterns replicate across domains: in coding, agents reassign or skip under-performing teammates; in research, they anchor plans in competent collaborators’ outputs; and in bargaining, milestones shift from unilateral moves to coordinated actions. Together, these results show that ETI improves coordination by supplying agents with actionable trait-based signals (see qualitative examples in Appendix~\ref{app:trait-examples_plans}–\ref{app:trait-examples_milestones}).

\paragraph{Cross-Model Differences in Trait Inference.} Quantitatively, across most scenarios, \textsc{Qwen} produces higher-variance ratings that span the full scale and better differentiate agents, whereas \textsc{GPT} ratings have low variance and are concentrated around positive values (see Appendix~\ref{app:trait_dist_marble}). Qualitatively, \textsc{Qwen} trait inference is more specific and diagnostic, while \textsc{GPT} tends to be generic and positively biased (examples in Appendix~\ref{app:trait-examples_traits}). The only exception is coding, where both models generate similarly objective trait inferences—likely because clear feedback (e.g., error messages) constrains ambiguity. Importantly, across all tasks, performance consistently aligns with these observed differences and similarities (Table~\ref{tab:overall_marble}).

Supporting this explanation of performance differences, we find a largely asymmetric pattern: \textsc{Qwen}-generated trait profiles mostly improve performance for both \textsc{Qwen} and \textsc{GPT} agents, whereas \textsc{GPT}-generated profiles often yield little or no benefit regardless of the acting model (Table \ref{tab:overall_marble}). This demonstrates that ETI’s gains stem from the informational quality of the profiles themselves, not from generic prompt restructuring, added context, or simply nudging agents to consider one another.

\section{Discussion}

Our study shows that equipping agents with ETI improves coordination and task performance. By providing actionable cues about others via trait representations, ETI's trait profiles help agents allocate work, anticipate adversaries, and reason with awareness of partners’ strengths and weaknesses, making agents not just better problem-solvers but more reliable collaborators across settings.

\paragraph{Relation to other social reasoning.} Existing approaches such as theory-of-mind or belief modeling focus on tracking transient states, such as current knowledge, beliefs, or intentions. ETI adds to these approaches by capturing more stable behavioral patterns, bridging short-term states and longer-term tendencies. Future work could investigate combining these approaches to build richer interaction models: knowing both what partners currently know, intend, or believe, and how they generally act (e.g., cooperative but not very competent), could enable more accurate prediction and planning.

\paragraph{Differences between model families.} We observe systematic differences across model families that surface directly in ETI profiles (e.g., Appendix Figure~\ref{fig:eti_example}). Pinpointing whether these differences arise from architecture, training data, or RLHF is beyond the scope of this work; however, prior findings on GPT-family positivity biases and sycophancy, often linked to RLHF \citep[e.g.,][]{fanous2025syceval}, are consistent with the patterns we see. Importantly, ETI’s value does not depend on identifying the precise cause: because these tendencies are measurable through profile generation, practitioners can use ETI as a lightweight diagnostic in their own domains to compare models and detect interaction-relevant biases before deployment.

\paragraph{Designing more integrated coordination.} Our results suggest that agents benefit when they reason explicitly about partners’ traits (e.g., ability, goal alignment), indicating that modeling and leveraging such information rather than optimizing only individual task reasoning is a promising direction for future MAS. A natural next step is to embed interdependent reasoning directly into models by training them in multi-agent environments with objectives that reward coordinated, partner aware reasoning and behavior. Such training could lead models to form latent representations of others’ characteristics—akin to how humans implicitly encode warmth and competence—rather than relying on predefined trait lists or templates. This shift would remove the need for manually specifying trait structures, allow agents to flexibly exploit whichever behavioral cues are relevant for a task, and reduce prompting overhead.

\section{Conclusion}
We introduced ETI, a lightweight, psychology-grounded framework that improves coordination in MAS through simple profiles that give agents structured awareness of others' traits. A key next step is to identify which model properties (e.g. architecture, training data, instruction tuning, or general reasoning abilities) enable effective trait inference, offering principled guidance for selecting and designing models for ETI and beyond. Achieving this could unlock socially intelligent multi-agent systems that reason about each other to self-organize, adapt, and coordinate with the fluidity and reliability of human teams.


\section*{Limitations}
Our findings demonstrate clear benefits of \emph{explicit trait inference} for multi-agent reasoning, yet several constraints of our study frame how broadly they can be interpreted. 

\paragraph{Model Scope.} Our experiments evaluate ETI on two small general-purpose LLMs (\textsc{GPT}-4o-mini and \textsc{Qwen}3-8B), in a prompt-based setup rather than fine-tuning. This restricted scope means the findings may not directly generalize to larger models, heterogeneous agent setups (e.g., mixtures of models), alternative architectures, or systems with task-specific training. Larger or fine-tuned models may exhibit different partner-modeling dynamics—potentially amplifying, reducing, or qualitatively changing the impact of ETI—depending on how implicit and explicit reasoning interact. Although preliminary tests (Appendix Section \ref{SOTA_Tests}) suggested that ETI can still be beneficial for larger models, future work should systematically examine these settings to assess ETI’s effectiveness across diverse agent configurations. 

\paragraph{Test Scope.} We evaluated ETI on four of the six \textsc{MultiAgentBench} scenarios (coding, research, negotiation, and social deduction). The two remaining scenarios were excluded because they offered limited additional qualitative insight relative to their engineering cost: database fixes overlap substantially with coding, and Minecraft, while collaborative like coding and research, is significantly harder to instrument and evaluate (e.g., linking trait profiles to the complex action space). Similarly, the two chosen economic games, while capturing distinct coordination challenges, necessarily represent only a narrow subset of possible dynamics. These choices allowed us to focus on settings that provide clear tests of ETI, but they also limit the scope of our conclusions to the selected tasks and interaction types.

\paragraph{Trait representation.} Our implementation used a fixed set of eight traits to instantiate the warmth and competence dimensions. 
While this made evaluation transparent and comparable across tasks, it does not adapt the granularity of inference to the needs of a specific domain or scenario.  
Future work could automatically select or learn task-relevant traits—rather than relying on a fixed set—or use latent trait representations (e.g., embedding-based) to capture more nuanced behavioral cues.

\paragraph{Bias, Calibration, and Adaptability} 
ETI inherits the idiosyncrasies of the underlying LLMs. For example, \textsc{GPT}-4o-mini showed sycophancy and positivity biases, producing more generic and uniformly positive trait inferences than \textsc{Qwen}3-8B, highlighting the need for model-specific calibration, more objective inference signals, or tuned prompting procedures.
Additionally, like any partner-modeling approach, ETI depends on trait inferences accurately reflecting others, as inaccurate (e.g., outdated) profiles can bias downstream decisions. Because ETI emphasizes stable, dominant patterns, it may potentially slow adaptation to abrupt partner shifts, degrading coordination, until sufficient new evidence accumulates (Appendix~\ref{app:adaptability}). At the same time, this conservatism is beneficial under noisy and ambiguous conditions: in our economic-game evaluations (Section~\ref{res:econ_games}; Table~\ref{tab:trait-inference}), profiles remain stable rather than overreacting to short-term deviations, and the adaptability experiment (Appendix~\ref{app:adaptability}) shows updates occur gradually as evidence builds.
Results from the Werewolf setting, where adversaries deliberately project false traits to gain and exploit trust, further indicate that the severity of this issue is context-dependent rather than universal. Future work should examine the extent of this limitation and whether lightweight strategies—such as sliding context windows or explicit change monitoring—can mitigate it.
Finally, trait inference—like any human social evaluation—can potentially reflect and amplify social biases. In real-world deployments involving demographic or social cues, ETI could reinforce stereotypes or unfair expectations. Future work should investigate safeguards and thoughtful prompt design to mitigate these risks in practice.

\paragraph{Scalability and Efficiency.} 
ETI is lightweight compared to agents’ core operations, since trait inference focuses on generating short profiles. However, by changing how agents plan, act, and communicate to improve coordination, ETI can indirectly increase total token usage—for example, by prompting more inter-agent communication to align on strategies. In our experiments, communication-heavy scenarios showed larger relative token increases; however, absolute usage remained moderate, so added cost and latency stayed low for the tested models. Appendix Table~\ref{tab:eti_cost_benefit} summarizes this trade-off: ETI can approach a $\sim$2$\times$ token increase in the heaviest-communication settings, yet the incremental cost per run is still small (roughly \$0.01--\$0.13 at GPT-4o/5-priced rates). Nevertheless, these behavioral shifts highlight the need to examine how ETI scales to large-scale deployments or more resource-constrained settings, and to explore strategies for controlling overhead in future work.  Several mitigation strategies are available. First, ETI calls can be throttled via sparse updates or truncated histories; our experiments in Appendix~\ref{appendix: warmup} and \ref{appendix: generalizability} show that a small number of calibration updates, or reusing profiles across tasks, retains most of the performance gains. Second, ETI-induced communication can be capped (e.g., maximum coordination rounds, verification calls, or total turns). Notably, ETI still improves outcomes in structurally constrained settings (Werewolf, Coding-Tree, where game/task design fixes the number of actions, length of output, etc), indicating that coordination benefits persist even under tight generation and communication budgets.

\section*{Acknowledgments}
We thank the AWS Agentic AI Labs team for their valuable feedback and support. We are especially grateful to Xiaofei Ma for his guidance and support throughout this project.

\bibliography{custom}

\clearpage
\appendix

\pagenumbering{roman}
\renewcommand{\thefigure}{A\arabic{figure}}
\renewcommand{\thetable}{A\arabic{table}}
\setcounter{figure}{0}
\setcounter{table}{0}

\section{Appendix}
\label{sec:appendix}

\subsection{Trait Definitions and Trait Inference Prompts}
\label{app:trait_definitions_prompts}

\subsubsection{Trait Definitions}
Table~\ref{tab:traits_definitions} provides the full definitions of the eight traits used to represent warmth and competence, along with the prompt templates. These definitions anchor trait judgments in concrete behavioral criteria relevant to agent interactions, following the framework outlined in the Methods.

\begin{table*}[htp]
\centering
\resizebox{0.8\textwidth}{!}{%
\begin{tabular}{p{0.18\linewidth}p{0.18\linewidth}p{0.58\linewidth}}
\toprule
\textbf{Category} & \textbf{Trait} & \textbf{Definition} \\
\midrule
Competence & Execution Ability & Extent to which the agent successfully carried out its chosen actions, regardless of whether the choices were optimal. \\
           & Reliability & Consistency in performance quality and strategy across rounds. \\
           & Adaptability & Ability to adjust approach in response to failures or changes in the environment/partner behavior. \\
           & Efficiency & Achieving quality outcomes with minimal attempts, effort, or wasted resources. \\
\midrule
Warmth & Goal Alignment & Whether the agent attempted choices serving the shared objective, regardless of success. \\
       & Collaboration & Efforts to coordinate actions or support mutual benefit. \\
       & Trustworthiness & Keeping commitments, acting honestly, and avoiding deception. \\
       & Maliciousness & Intentional sabotage of collective progress, beyond self-interest (e.g., harming others even at self-cost). \\
\bottomrule
\end{tabular}%
}
\caption{Trait categories, names, and operational definitions used for prompting the agents.}
\label{tab:traits_definitions}
\end{table*}

\subsubsection{Trait Inference Generation}
After each interaction, we elicited updated trait inferences for each interaction partner of a given agent. 
The model was instructed to review the \emph{scenario context}---consisting of (a) a general task description (e.g., payoff matrix in an economic game, or high-level MAS goal with requirements in MultiAgentBench), and (b) the interaction history (e.g., sequences of actions and payoffs in games, or observations like code edits, error messages, and communication logs in MultiAgentBench). 
Based on this input and the fixed set of eight trait definitions, shown in Table \ref{tab:traits_definitions}, the model produced a structured JSON profile with a 1--7 rating for each trait and a short evidence summary. See an example of the trait inference prompt in Figure \ref{templ:trait_inference}.

\begin{figure*}[!t]
\begin{mdframed}[backgroundcolor=gray!07,linecolor=gray!40,roundcorner=2pt]
\small
\begin{verbatim}
System: You are an expert at inferring stable traits from interactions.

Inputs:
  {scenario_description}   # Game rules, MAS objective, requirements
  {interaction_history}    # Actions, payoffs, edits, errors, messages
  {trait_definitions}      # 8 predefined traits

Task:
  - Analyze the interaction history
  - Identify stable behavioral patterns
  - Assign ratings (1--7 or N/A) for each trait
  - Provide 2-3 sentence evidence per trait

Output:
  {json_template}
\end{verbatim}
\end{mdframed}
\caption{Template for trait inference prompting.}
\label{templ:trait_inference}
\end{figure*}

\subsubsection{Trait Profile Injection}
At decision time, we appended additional content to the original prompts for acting, planning, and communicating. 
This consisted of a short preamble indicating that information about relevant collaborators was available (e.g., ``Here is some information about your collaborators''), followed by the most recent trait profiles of those agents, and finally a brief instruction such as ``Use this information to make better decisions, coordinate better, and improve overall outcomes.'' See an example of the trait profile injection in Figure \ref{templ:trait_injection}.

\begin{figure*}[!t]
\begin{mdframed}[backgroundcolor=gray!07,linecolor=gray!40,roundcorner=2pt]
\small
\begin{verbatim}
You are {agent_id}: {self_profile}

Task:
  {task_description}

Context:
  - Other agents available: {agent_descriptions}
  - Your memory: {agent_memory}

Action Instruction:
{Instructions to generate output for next action}

Trait Profiles:
  {partner_id_1}: {trait_profile_1}
  {partner_id_2}: {trait_profile_2}
  ...
  Use this information to make better decisions, 
  coordinate better, and improve overall outcomes.
\end{verbatim}
\end{mdframed}
\caption{Template for trait profile injection into act/plan/communication prompts.}
\label{templ:trait_injection}
\end{figure*}

\subsection{Economic Games}
\label{app:econ_game_section}

\subsubsection{LLM Inference Parameters}
For all experiments, we use the default inference settings for \textsc{Qwen}3-8B recommended in the documentation: we set temperature to 0.6, top-p to 0.95, top-k to 20, and MinP to 0.

\subsubsection{Game Mechanism.} Each game proceeds in repeated rounds where two agents simultaneously select one of two actions. Payoffs depend on the joint action and whether each action succeeds, which is determined by the agent’s competence level. At each round, agents observe the full history of previous actions and outcomes, then decide their next action based on this context. After both agents act, payoffs are revealed and appended to the shared interaction history. This loop repeats for 50 rounds.

\subsubsection{Pay-off Matrices}

Table~\ref{tab:stag_hunt_payoff} and Table~\ref{tab:pd_payoff} show the payoff matrices for the Stag Hunt and Prisoner’s Dilemma used in our experiments. Rows correspond to the agent’s actions, and columns to the opponent’s actions, with outcomes split by whether execution succeeded or failed. Competence is reflected as failures: when an agent is low-competence, intended actions may misfire, producing zero or altered payoffs. For instance, a failed attempt to defect may collapse into mutual punishment or even backfire on the defector. This design makes competence observable from outcomes, ensuring that optimal strategies depend on both intent (cooperation vs.\ defection) and capability.

\begin{table}[t!]
\centering
\resizebox{1.0\columnwidth}{!}{%
\begin{tabular}{l l cc cc}
\toprule
 & & \multicolumn{4}{c}{\textbf{Opponent Action}} \\
\cmidrule(lr){3-6}
 & & \multicolumn{2}{c}{Stag} & \multicolumn{2}{c}{Hare} \\
\cmidrule(lr){3-4} \cmidrule(lr){5-6}
\multirow{-3}{*}{\textbf{Agent Action}} & 
\multirow{-2}{*}{ } & Success & Failure & Success & Failure \\
\midrule
Stag & & (5,5) & (0,0) & (0,2) & (0,0) \\
Hare & & (2,0) & (2,0) & (2,2) & (2,0) \\
\bottomrule
\end{tabular}%
}
\caption{Payoff matrix for the Stag Hunt game. Each cell shows payoffs as (agent, opponent), 
with success/failure indicating whether the action succeeded.}
\label{tab:stag_hunt_payoff}
\end{table}

\begin{table}[t]
\centering
\resizebox{1.0\columnwidth}{!}{%
\begin{tabular}{l l cc cc}
\toprule
 & & \multicolumn{4}{c}{\textbf{Opponent Action}} \\
\cmidrule(lr){3-6}
 & & \multicolumn{2}{c}{Silent} & \multicolumn{2}{c}{Testify} \\
\cmidrule(lr){3-4} \cmidrule(lr){5-6}
\multirow{-3}{*}{\textbf{Agent Action}} & 
\multirow{-2}{*}{ } & Success & Failure & Success & Failure \\
\midrule
Testify & & (0,-10) & (-5,-5) & (-2,-2) & (-5,-10) \\
Silent    & & (-0.5,-.5) & (-0.5,-5) & (-10,0) & (-0.5,-5) \\
\bottomrule
\end{tabular}%
}
\caption{Payoff matrix for the Prisoner’s Dilemma. Each cell shows payoffs as (agent, opponent), 
with success/failure indicating whether the action succeeded.}
\label{tab:pd_payoff}
\end{table}

\begin{figure*}[!t]
\begin{mdframed}[backgroundcolor=gray!07,linecolor=gray!40,roundcorner=2pt]
\small
\begin{verbatim}
Inferred Trait Profile for Agent X:

"competence": {
  "execution_ability": {
    "rating": 5,
    "evidence": "Agent X consistently chooses Hare and almost always succeeds, 
                indicating strong execution ability in solo hunts."
  },
  "reliability": {
    "rating": 3,
    "evidence": "Performance shows variability, with occasional failures and noise 
                despite a dominant Hare strategy."
  },
  "adaptability": {
    "rating": 2,
    "evidence": "Agent X rarely changes strategy, continuing to choose Hare 
                even after repeated coordination failures on Stag."
  },
  "efficiency": {
    "rating": 6,
    "evidence": "Focuses on Hare, which yields steady payoffs with minimal risk 
                or coordination effort, suggesting efficient resource use."
  }
},

"warmth": {
  "goal_alignment": {
    "rating": 2,
    "evidence": "Seldom attempts Stag, showing little effort to pursue mutual benefit."
  },
  "collaboration": {
    "rating": 1,
    "evidence": "No evidence of coordinating with the partner on Stag choices, 
                suggesting absence of collaborative intent."
  },
  "trustworthiness": {
    "rating": 5,
    "evidence": "Acts consistently without deception or sabotage, 
                even if self-interested."
  },
  "maliciousness": {
    "rating": 1,
    "evidence": "No indication of intentional harm; defection stems 
                from self-interest, not sabotage."
  }
}
\end{verbatim}
\end{mdframed}
\caption{Trait profile during a game of Stag Hunt against an uncooperative but competent opponent, which the trait profile reflects. Note that profile distinguishes between different kinds of uncooperative (malicious vs untrustworthy vs uncooperative) and provides concrete reasoning.}
\label{templ:trait_profile}
\end{figure*}

\subsubsection{Example Trait Profiles}
\label{app:econ_trait_example}

We illustrate the trait inference process on a Stag Hunt trajectory with a \emph{competent but uncooperative} opponent in the noisy condition (15\% random actions/outcomes). The opponent defected and executed actions reliably, with occasional stochastic deviations. See the resulting trait profile in Figure \ref{templ:trait_profile}.

\subsubsection{Prisoner Dilemma: Payoff Optimality }
\begin{figure}[t]
  \includegraphics[width=.9\columnwidth]{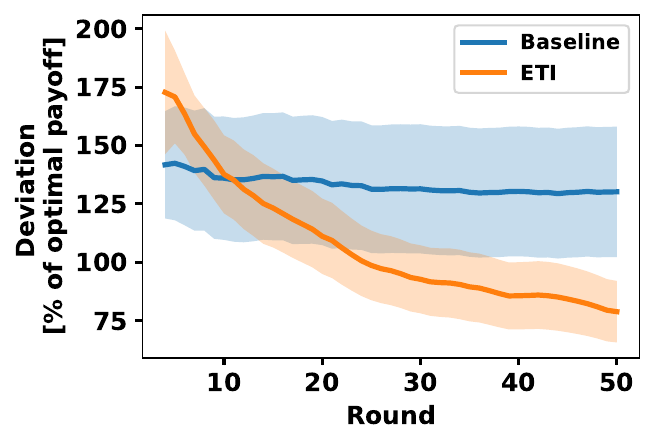}
  \caption{Relative deviation from optimal payoff across rounds (Prisoner Dilemma). Lower is better. Shading represents 95-\% Confidence Interval}
  \label{fig:deviation_prisoner}
\end{figure}

As shown in Figure~\ref{fig:deviation_prisoner} and mirroring Figure~\ref{fig:deviation_staghunt}, ETI agents achieve markedly lower relative deviation from the optimal payoff in the Prisoner’s Dilemma. While both modes improve over time, ETI consistently outperforms the baseline, highlighting its advantage in approaching payoff-optimal strategies.

\subsubsection{Choice Optimality Across Rounds}
\begin{figure}[t]
  \includegraphics[width=.9\columnwidth]{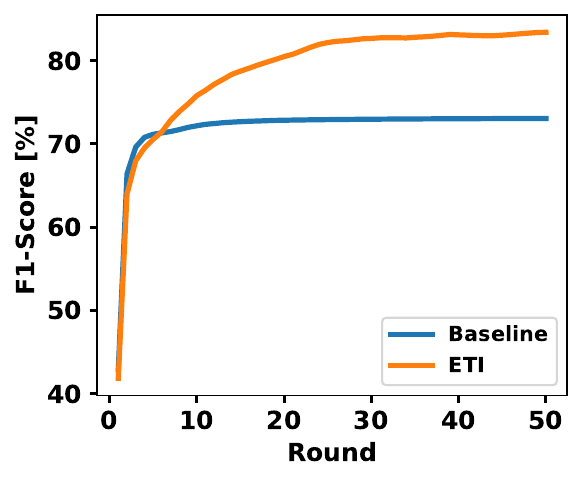}
  \caption{Optimal decisions over rounds  in main experiment (Prisoner Dilemma)}
  \label{fig:cumulative_f1_IPD}
\end{figure}

\begin{figure}[t]
  \includegraphics[width=.9\columnwidth]{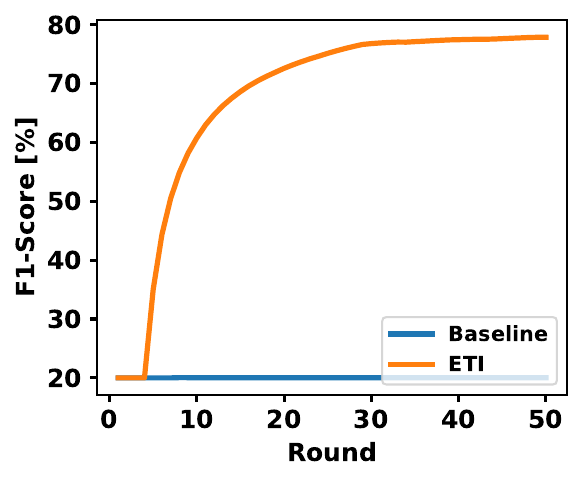}
  \caption{Optimal decisions over rounds  in main experiment (Stag Hunt)}
  \label{fig:cumulative_f1_SH}
\end{figure}

In Figure \ref{fig:cumulative_f1_IPD} and Figure \ref{fig:cumulative_f1_SH}, we show the F1 score of agents making the game-theoretically optimal decision across rounds in the Prisoner’s Dilemma and Stag Hunt. ETI yields consistently higher F1 scores and consistently reaches strong performance (F1 > 0.8), indicating that trait-informed reasoning enhances the model’s ability to identify and sustain optimal strategic choices over repeated interactions.

\begin{table}[t]
\centering
\resizebox{0.95\columnwidth}{!}{%
\setlength{\tabcolsep}{6pt}       
\renewcommand{\arraystretch}{1.0} 
\begin{tabular}{@{}llccc@{}} 
\toprule
Game & Trait & Baseline & In-task & Warm-up \\
\midrule
PD & Comp. & 0.69 & 0.88* & 0.83* \\
    & Coop. & 0.43 & 0.74* & \textbf{0.78*} \\
\addlinespace
SH  & Comp. & 0.85 & 0.88* & \textbf{0.93*} \\
    & Coop. & 0.52 & \textbf{0.79*} & 0.70* \\
\bottomrule
\end{tabular}%
}
\caption{Trait inference accuracy (F1) under three modes: \textbf{Baseline} (no trait inference), \textbf{In-task} (profiles inferred and updated continuously), and \textbf{Warm-up} (profiles calibrated in a prior interaction and then fixed). Bold indicates the best performance per trait and game. Asterisks (*) mark statistically significant improvements over baseline ($p < 0.05$).}
\label{tab:trait-inference_warmup}
\end{table}

\subsubsection{Pre-calibration of Trait Inference}
\label{appendix: warmup}

Our main experiments rely on continuous trait inference: agents update trait profiles of their partners after each round, enabling rapid adaptation but incurring token and compute costs. Conceptually, however, traits reflect relatively stable behavioral patterns. This raises the question of whether traits need to be inferred continuously, or whether they can be \emph{pre-calibrated} from a set of past interactions and then kept fixed for future use, similar to how human evaluations stabilize after an initial observation period.

\paragraph{Design.}  
We ran a two-stage setup. First, agents played 25 rounds against a scripted opponent, and we generated a trait profile from this interaction. Then, the agent played another game against a scripted opponent with the same latent traits (e.g., cooperative but incompetent) but started with this frozen trait profile. Importantly, the agents did not have access to past trajectories or any other information about the previous run except the pre-calibrated trait profiles. During the test game, profiles were not updated. We compare the performance against agents 

\paragraph{Results.}  
See Table \ref{tab:trait-inference_warmup} for a comparison of agents without trait inference (baseline), with continuously updated trait inference (in-task), and with pre-calibrated trait inference (warm-up). Pre-calibrated static profiles substantially improve trait inference accuracy and decision quality relative to the baseline (no trait inference). Moreover, performance is largely comparable to that of continuously updated profiles. For example, in detecting latent competence and cooperation, pre-calibrated profiles achieve nearly the same accuracy as in-task updated profiles. Similarly, pre-calibration yields comparably optimal downstream decisions: agents achieve high choice accuracy (Figs.~\ref{fig:cumulative_f1_IPD}, \ref{fig:cumulative_f1_SH}) and reduced payoff deviation, on par with continuously updated profiles Importantly, pre-calibration provides a marked advantage in the early rounds, when continuous profiles are still inaccurate due to limited observations (often requiring $\sim$25 rounds to converge).

\paragraph{Implications.}  
Results show that continuous updates are not strictly necessary for ETI to improve decision making. Trait profiles can be generalized from prior interactions with the same agent (at least within the same task), and used to bootstrap coordination in future games. This implies that MAS systems could cache trait profiles from earlier encounters and reuse them in later tasks, reducing the need for continuous monitoring and saving tokens, compute, and time.

\begin{figure}[t]
  \includegraphics[width=\columnwidth]{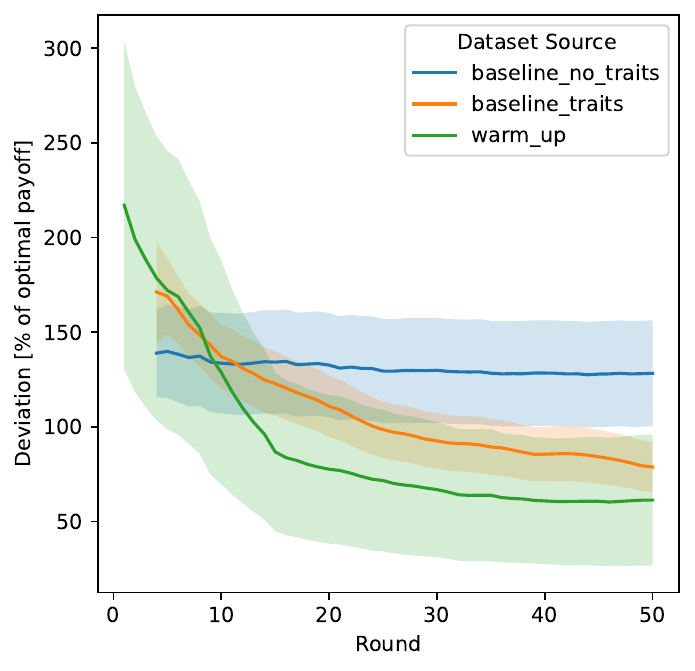}
  \caption{Relative deviation from optimal payoff across rounds for Prisoner Dilemma. Shading represents 95-\% Confidence Interval. Lower is better.}
  \label{fig:deviation_stag_warm}
\end{figure}

\begin{figure}[t]
  \includegraphics[width=\columnwidth]{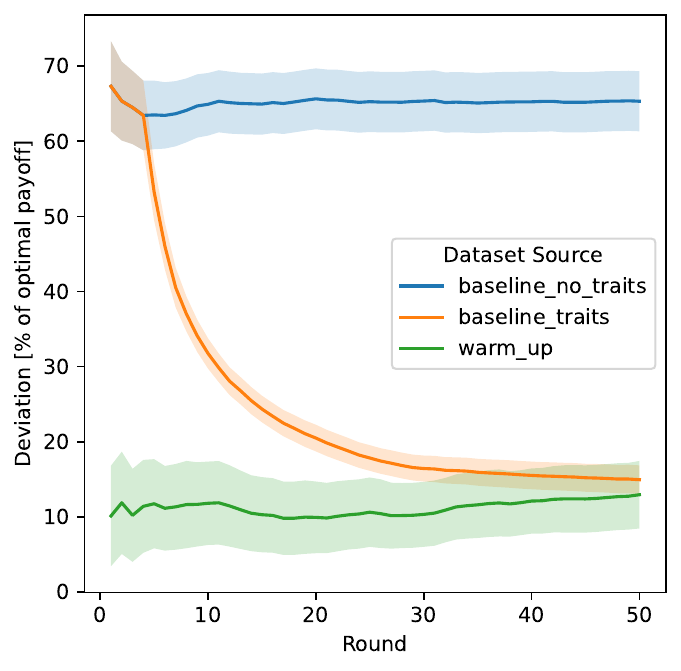}
  \caption{Relative deviation from optimal payoff across rounds for Stag Hunt. Shading represents 95-\% Confidence Interval. Lower is better.}
  \label{fig:deviation_prisoner_warm}
\end{figure}

\begin{table}[htp]
\centering
\resizebox{0.95\columnwidth}{!}{%
\setlength{\tabcolsep}{6pt}       
\renewcommand{\arraystretch}{1.0} 
\begin{tabular}{@{}llccc@{}} 
\toprule
Game & Trait & Baseline & In-task & Cross-Task \\
\midrule
PD & Comp. & 0.69 & 0.91* & \textbf{0.97*} \\
    & Coop. & 0.43 & \textbf{0.72*} & 0.70* \\
\addlinespace
SH  & Comp. & 0.85 & \textbf{0.85} & 0.79 \\
    & Coop. & 0.52 & \textbf{0.77*} & 0.59* \\
\bottomrule
\end{tabular}%
}
\caption{Trait inference accuracy (F1) under three modes: \textbf{Baseline} (no trait inference), \textbf{In-task} (profiles inferred and updated within the same task), and \textbf{Cross-task} (profiles inferred in a different task and kept fixed). Bold indicates the best performance per trait and game. Asterisks (*) mark statistically significant improvements over baseline ($p < 0.05$).}
\label{tab:trait-inference_generalizability}
\end{table}

\subsubsection{Cross-Task Generalization of Trait Inference}
\label{appendix: generalizability}

A key assumption in our framework is that traits such as competence and cooperation reflect relatively stable behavioral dispositions. If so, trait profiles learned in one context should transfer to another, reducing the need to re-estimate traits in every new interaction \textbf{if} (a) the underlying skills are relevant across tasks, and (b) agents can correctly apply trait information even when behavioral manifestations differ across tasks (e.g., the cooperative action in prisoner dilemma is different from stag hunt), and c) agents can correctly integrate the cross-task profile to find the optimal response strategy in the new task (e.g., an incompetent partner in prisoner dilemma requires a different response than in stag hunt).

\paragraph{Design.}  
We test cross-task generalization by letting agents first complete one game against a scripted opponent (e.g., competent but uncooperative), from which a final trait profile is generated and stored. The agent then plays the \emph{other} game (Stag Hunt $\to$ Prisoner’s Dilemma or vice versa) against the same opponent. The transferred profile is provided at the start of the new game, but no in-task updates are performed, ensuring that only cross-task information is available.

\begin{figure}[t]
  \includegraphics[width=\columnwidth]{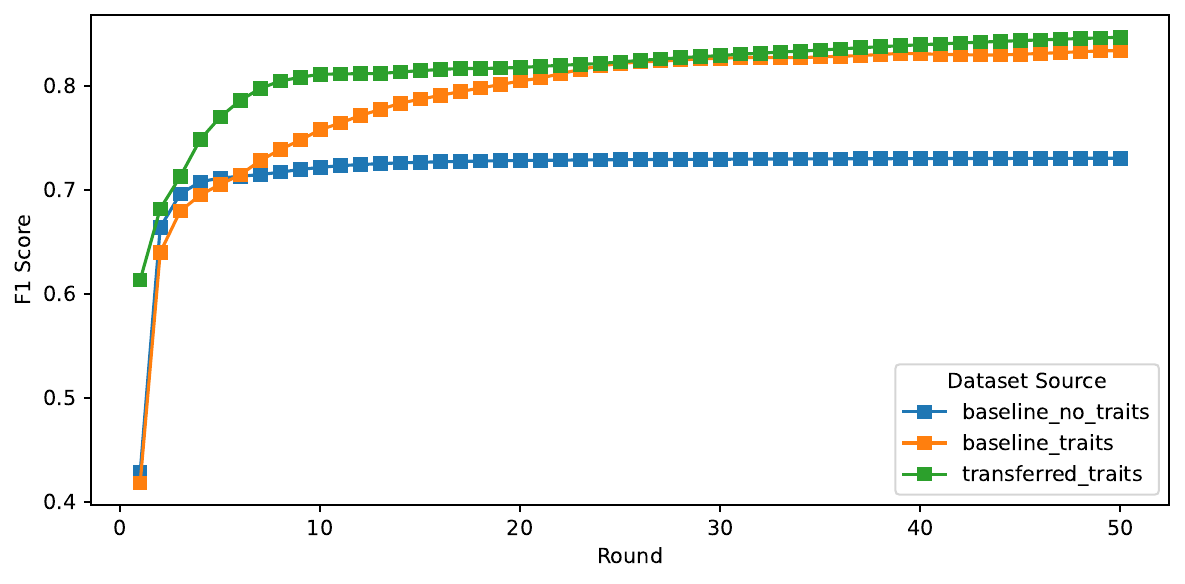}
  \caption{Optimal decisions over rounds (Prisoner Dilemma) in generalization experiment.}
  \label{fig:cumulative_f1_IPD_generalizability}
\end{figure}

\begin{figure}[t]
  \includegraphics[width=\columnwidth]{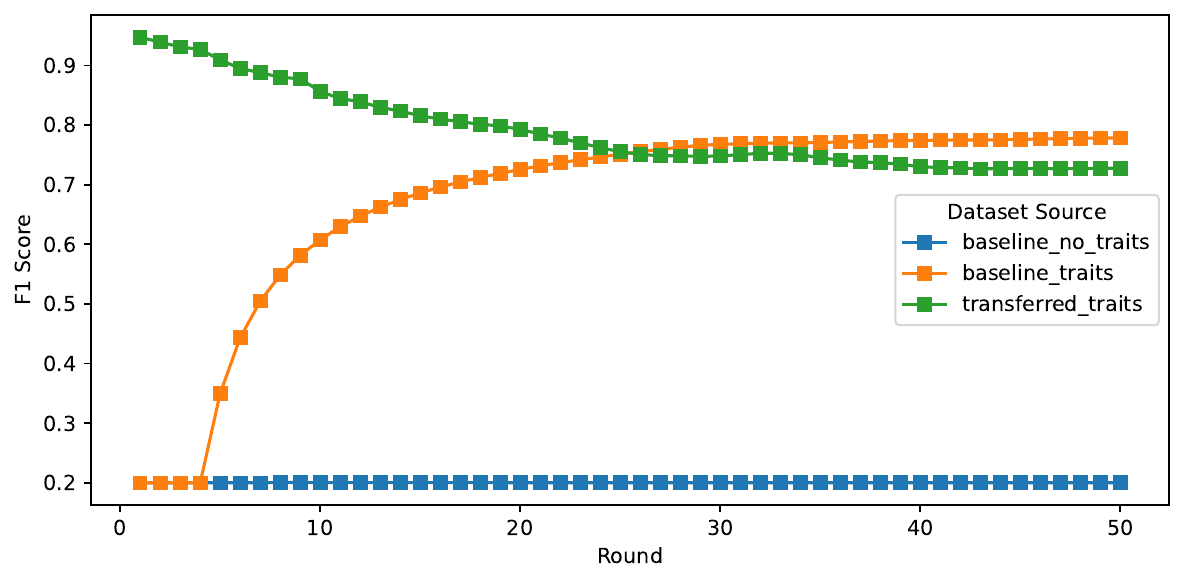}
  \caption{Optimal decisions over rounds (Stag Hunt) in generalization experiment.}
  \label{fig:cumulative_f1_SH_generalizability}
\end{figure}

\begin{figure}[t]
  \includegraphics[width=\columnwidth]{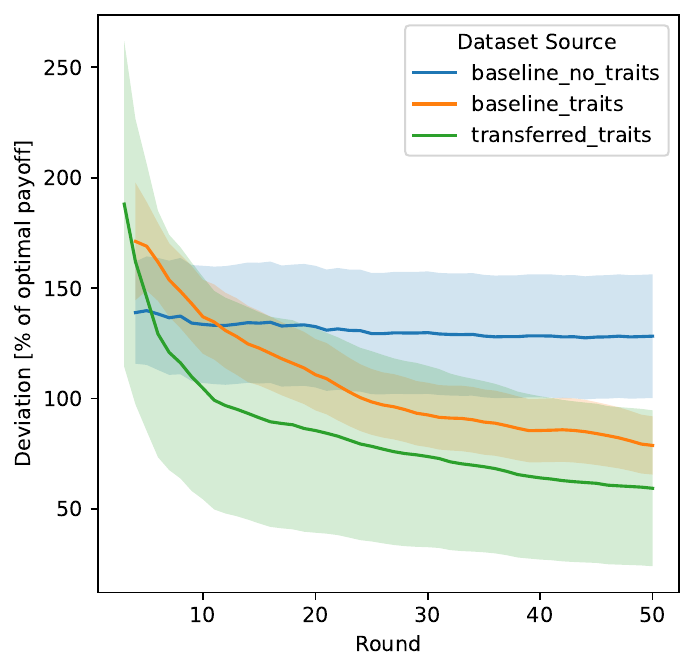}
  \caption{Relative deviation from optimal payoff across rounds (Prisoner Dilemma). Shading represents 95-\% Confidence Interval. Lower is better.}
  \label{fig:deviation_stag_generalizability}
\end{figure}

\begin{figure}[t]
  \includegraphics[width=\columnwidth]{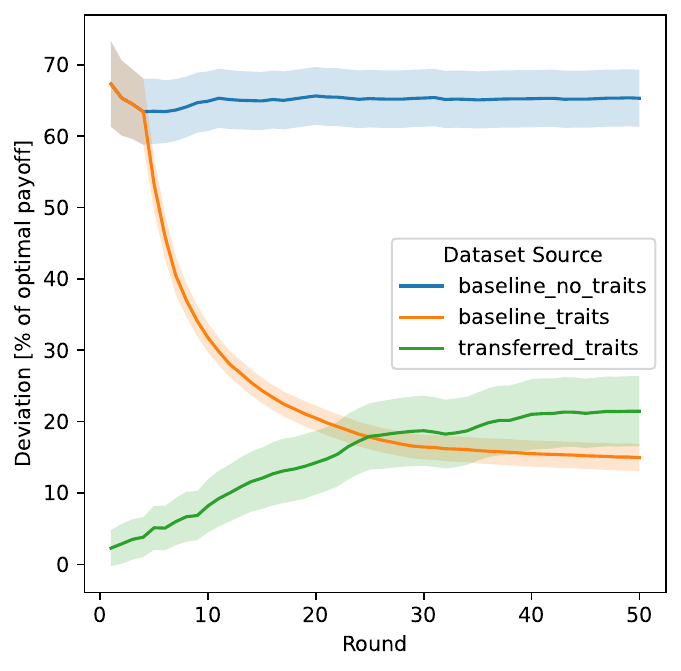}
  \caption{Relative deviation from optimal payoff across rounds (Stag Hunt). Shading represents 95-\% Confidence Interval. Lower is better.}
  \label{fig:deviation_prisoner_generalizability}
\end{figure}

\paragraph{Results.}  
See Table \ref{tab:trait-inference_generalizability} for a comparison of agents without trait inference (baseline), with in-task and cross-task trait inference. We find that cross-task profiles substantially improve both trait inference accuracy and decision quality relative to the baseline, and perform nearly as well as in-task profiles in most conditions. Transferring competence inferences is particularly robust: F1 remains high in both directions. Cooperation transfer is more asymmetric: while Stag Hunt $\to$ Prisoner’s Dilemma works reliably, transferring from Prisoner’s Dilemma $\to$ Stag Hunt is weaker, though still better than baseline. This is consistent with our main results showing that cooperation inference is more challenging in the PD (F1 = 0.73) than in the SH (F1 = 0.81). Furthermore, investigating the choice optimality (Figures~\ref{fig:cumulative_f1_IPD_generalizability}, \ref{fig:cumulative_f1_SH_generalizability}) and payoffs across rounds (Figures~\ref{fig:deviation_prisoner_generalizability}, \ref{fig:deviation_stag_generalizability}) clearly shows that generalized trait inference performs as well as in-task trait inference long term and even provides initial advantages, likely due to calibration effects as shown in Section \ref{appendix: warmup}.

\paragraph{Implications.}  
These findings suggest that trait profiles generalize beyond the immediate context in which they were inferred. This opens the possibility of building long-term partner models in MAS—where profiles generated in one setting (e.g., debugging code) can inform interactions in another (e.g., research collaboration). While our results demonstrate transfer across structurally similar economic games, further work is needed to test how far such generalization extends to tasks with more diverse structures and interaction patterns.


\subsubsection{Effects on Adaptability}
\label{app:adaptability}
ETI is designed to capture stable partner traits, which may slow adaptation when behavior changes abruptly (e.g., model updates or deception) and new evidence must override earlier inferences. We test ETI’s responsiveness to such shifts by examining how quickly trait profiles update when a partner changes (e.g., competent to incompetent).

\paragraph{Design.}
Agents play a 50-round game against a scripted opponent with a fixed trait profile for the first 25 rounds; at mid-point (round 25) the opponent’s traits switch (e.g., cooperative$\to$uncooperative or competent$\to$incompetent). We evaluate post-shift performance: (i) trait inference accuracy for the new latent traits, (ii) choice optimality, and (iii) payoff deviation from the optimal strategy.

\paragraph{Results.}
\textit{Trait inference accuracy.}
Table \ref{tab:trait-inference_adaptability} compares agents without trait inference (baseline) to those using ETI after a mid-game trait shift. ETI agents show reduced accuracy at detecting the new traits compared to no-shift conditions, indicating inertia toward prior dominant patterns. In PD, post-shift accuracy remains above baseline but below or equal to no-shift performance. In SH, ETI is weaker post-shift for competence and roughly on par with baseline for cooperation, suggesting that emphasizing stable patterns can lead to underweighting recent evidence when behavior changes abruptly.

\begin{table}[t]
\centering
\setlength{\tabcolsep}{6pt}       
\renewcommand{\arraystretch}{1.0} 
\begin{tabular}{@{}llcc@{}}
\toprule
Game & Trait & Baseline & ETI \\
\midrule
PD & Comp. & 0.10 & \textbf{0.43*} \\
    & Coop. & 0.28 & \textbf{0.45*}\\
\addlinespace
SH  & Comp. & \textbf{0.71*} & 0.59  \\
    & Coop. & \textbf{0.46} & 0.45 \\
\bottomrule
\end{tabular}
\caption{Trait inference accuracy (F1) after abrupt behavior changes. Bold indicates best performance per trait and game. Asterisks (*) mark significant differences ($p < 0.05$).}
\label{tab:trait-inference_adaptability}
\end{table}

\begin{figure}[t]
  \includegraphics[width=\columnwidth]{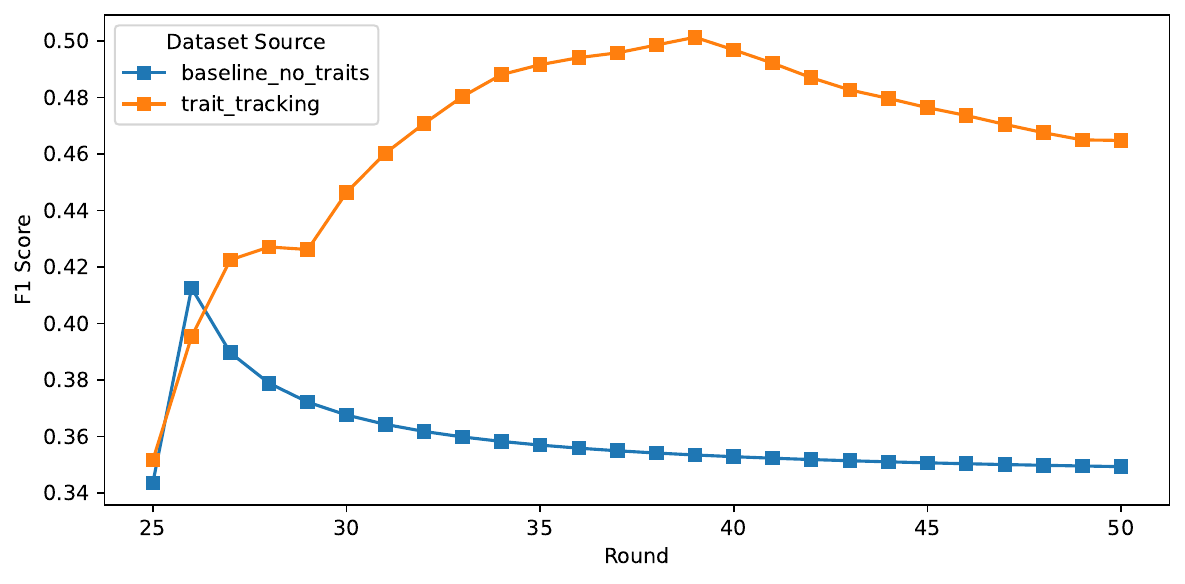}
  \caption{Optimal decisions over rounds (Prisoner Dilemma) in adaptability experiment.}
  \label{fig:cumulative_f1_IPD_adaptability}
\end{figure}

\begin{figure}[t]
  \includegraphics[width=\columnwidth]{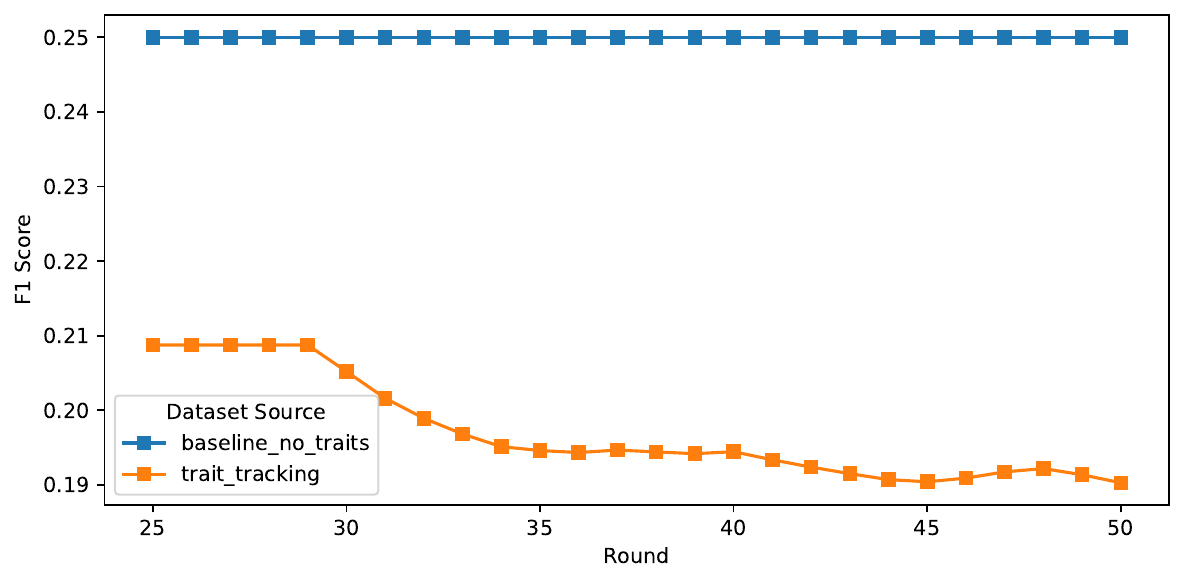}
  \caption{Optimal decisions over rounds (Stag Hunt)  in adaptability experiment.}
  \label{fig:cumulative_f1_SH_adaptability}
\end{figure}

\textit{Choice optimality.}
Consistent with accuracy, ETI agents are slower to adopt the optimal response after the shift. In PD, optimal-choice F1 improves over time and exceeds baseline but lags behind the no-shift trajectory; in SH, ETI's slow adaptation yields lower post-shift optimal-choice F1 than baseline (Figs.~\ref{fig:cumulative_f1_IPD_adaptability}, \ref{fig:cumulative_f1_SH_adaptability}).

\begin{figure}[t]
  \includegraphics[width=\columnwidth]{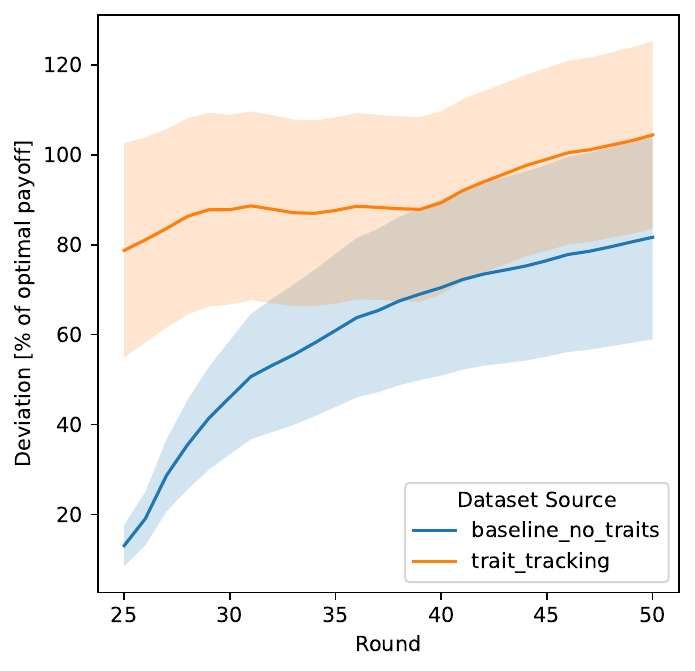}
  \caption{Relative deviation from optimal payoff across rounds (Prisoner Dilemma). Shading represents 95-\% Confidence Interval. Lower is better.}
  \label{fig:deviation_stag_adaptability}
\end{figure}

\begin{figure}[t]
  \includegraphics[width=\columnwidth]{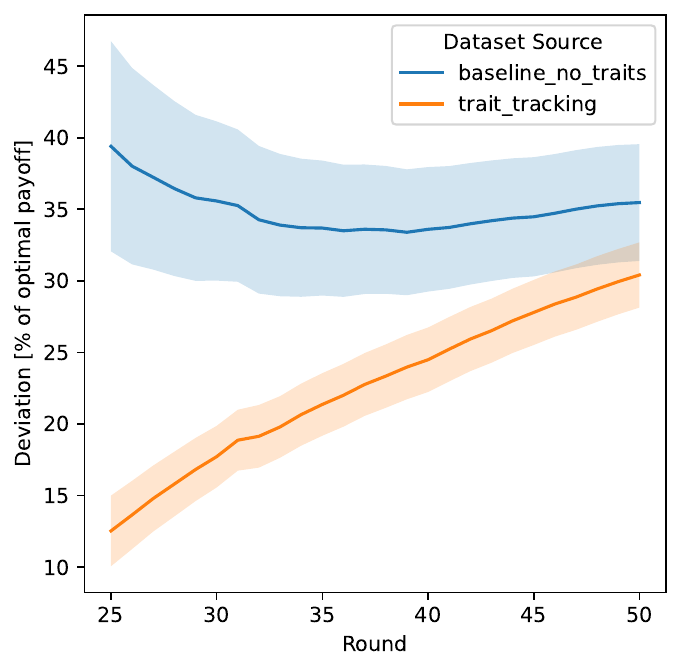}
  \caption{Relative deviation from optimal payoff across rounds (Stag Hunt). Shading represents 95-\% Confidence Interval. Lower is better.}
  \label{fig:deviation_prisoner_adaptability}
\end{figure}

\textit{Payoffs.}
Post-shift, relative deviation from optimal payoffs increases for ETI agents in both games, with SH showing the largest gap (Figs.~\ref{fig:deviation_stag_adaptability}, \ref{fig:deviation_prisoner_adaptability}). Deviation declines as more post-shift evidence accumulates, but convergence is slower than in no-shift conditions.

\textit{Qualitative observation.}
Transcript inspection (not shown) reveals inertia effects: after a cooperative$\to$uncooperative switch, agents often continue proposing cooperative plans (e.g., Stag) for several rounds, justifying these choices based on earlier evidence before updating to the new pattern.


\paragraph{Implications.}
ETI can reduce adaptability to abrupt behavioral changes, likely because our implementation emphasizes dominant historical patterns. However, (a) mid-game shifts also challenge baseline agents, (b) ETI still outperforms baseline in several post-shift PD settings, and (c) we did not equip agents with any change-aware mechanisms. This points to simple extensions—e.g., recency weighting, sliding context windows, or explicit change prompts—that could preserve ETI’s benefits while improving responsiveness to changing partners.


\subsection{MultiAgentBench}

In our experiments on MultiAgentBench, we generally adhered to the example setup provided in the benchmark repository and used the default parameter settings (e.g., temperature, top-k, top-p, and random seed). For Qwen models, however, we followed the documentation’s recommended adjustments for tool-use scenarios: specifically, we set temperature to 0.6, top-p to 0.95, top-k to 20, and MinP to 0. Outside of these cases, all parameters were kept consistent with the benchmark defaults, which were originally specified for \textsc{GPT}-4o and \textsc{GPT}-4o-mini.

\subsubsection{Additional Metrics}
\label{app:add_metrics}
In the following, we present supplementary metrics beyond the main \emph{Task Score} and \emph{Coordination Score} reported in the paper, across all MultiAgentBench scenarios. All additional metrics are derived from the benchmark framework (e.g., communication score is a subcomponent of coordination) and follow \citet{zhu2025multiagentbench}, unless explicitly noted.

Metrics are a mix of (i) objective measurements (e.g., token usage, communication counts, tasks completed, games won), 
(ii) predefined auto-evaluated milestones (e.g., successful execution of specific actions), 
and (iii) LLM-judge based rubric evaluations (e.g., communication quality, effective reasoning, and planning). 
Following the benchmark setup, we use \textsc{GPT}-4o as the judge model with the same rubrics and prompts.

For simplicity, across all tables and results we use the shorthand \textsc{GPT} to refer to \textsc{GPT}-4o-mini and \textsc{Qwen} to refer to \textsc{Qwen}3-8B, both for agent models and for trait inference models.

\begin{table}[!thp]
\centering
\begin{tabular}{>{\arraybackslash}p{1.0cm} >{\arraybackslash}p{1cm} >{\arraybackslash}p{2.5cm} c}
\toprule
\textbf{Agent Model} & \textbf{ETI Model} & \textbf{Metric} & \textbf{Mean} \\
\midrule
\rowcolor{gray!15} \textsc{GPT}  & none        & Villager WR & 26.00 \\
\textsc{GPT}  & \textsc{GPT} & Villager WR & 32.00* \\
\textsc{GPT}  & \textsc{Qwen}    & Villager WR & 44.00* \\
\rowcolor{gray!15} \textsc{Qwen} & none        & Villager WR & 54.00 \\
\textsc{Qwen} & \textsc{Qwen}    & Villager WR & \textbf{64.00}* \\
\textsc{Qwen} & \textsc{GPT} & Villager WR & 60.00* \\
\midrule
\rowcolor{gray!15} \textsc{GPT}  & none        & Completion   & 29.80 \\
\textsc{GPT}  & \textsc{GPT} & Completion   & 29.25 \\
\textsc{GPT}  & \textsc{Qwen}    & Completion   & 36.03* \\
\rowcolor{gray!15} \textsc{Qwen} & none        & Completion   & 39.24 \\
\textsc{Qwen} & \textsc{Qwen}    & Completion   & \textbf{51.09}* \\
\textsc{Qwen} & \textsc{GPT} & Completion   & 44.66* \\
\midrule
\rowcolor{gray!15} \textsc{GPT}  & none        & Result Score      & 0.22 \\
\textsc{GPT}  & \textsc{GPT} & Result Score      & 0.42 \\
\textsc{GPT}  & \textsc{Qwen}    & Result Score      & 0.70* \\
\rowcolor{gray!15} \textsc{Qwen} & none        & Result Score      & 1.32 \\
\textsc{Qwen} & \textsc{Qwen}    & Result Score      & \textbf{2.22}* \\
\textsc{Qwen} & \textsc{GPT} & Result Score      & 1.48* \\
\midrule
\rowcolor{gray!15} \textsc{GPT}  & none        & Total Net Score   & -0.83 \\
\textsc{GPT}  & \textsc{GPT} & Total Net Score   & 0.01 \\
\textsc{GPT}  & \textsc{Qwen}    & Total Net Score   & 0.22 \\
\rowcolor{gray!15} \textsc{Qwen} & none        & Total Net Score   & 2.13 \\
\textsc{Qwen} & \textsc{Qwen}    & Total Net Score   & \textbf{5.50}* \\
\textsc{Qwen} & \textsc{GPT} & Total Net Score   & 4.08* \\
\bottomrule
\end{tabular}
\caption{Table shows the supplemental performance metrics for the werewolf scenario. ETI improves performance across both short- and long-term metrics. Asterisks (*) indicate significant improvement over baseline ($p<0.05$). Bold indicates best-performing run.}
\label{app:werewolf_perf_extra}
\end{table}

\paragraph{Werewolf}
For the Werewolf scenario various performance related scores were calculated (Table \ref{app:werewolf_perf_extra}). Note that the evaluations were focused on the villager agents consistent with the benchmark.

\begin{itemize}
    \item \textit{Daily Task Completion (Completion)}:  
    Average percentage of positive actions completed by villagers (e.g., voting out a werewolf) out of the maximum possible in each round. This expresses how well villager agents executed beneficial actions in the short-term (round-by-round basis).

    \item \textit{Total Net Score}:  
    Net difference in points accumulated by villagers vs.\ werewolves over the game. Points reflect successful actions toward each side’s goals (e.g., villagers get $+1$ for voting out a werewolf), capturing relative short-term performance (round-by-round basis).

    \item \textit{Result Score}:  
    Difference between surviving villagers and werewolves at game end, reflecting the strength of the win or loss. Larger positive values indicate dominant villager wins; larger negative values indicate dominant werewolf victories.

    \item \textit{Village Win Rate (WR)}:  
    Average percentage of games won by the villager faction, expressing long-term performance.
    
\end{itemize}

\begin{table}[!thp]
\centering
\begin{tabular}{>{\arraybackslash}p{1.25cm} >{\arraybackslash}p{1cm} >{\arraybackslash}p{1.25cm} c}
\toprule
\textbf{Agent Model} & \textbf{ETI Model} & \textbf{Metric} & \textbf{Mean} \\
\midrule
\rowcolor{gray!15} \textsc{GPT}  & none & Info    & 69.2 \\
\textsc{GPT}  & \textsc{GPT}  & Info    & 73.6* \\
\textsc{GPT}  & \textsc{Qwen} & Info    & 74.4* \\
\rowcolor{gray!15} \textsc{Qwen} & none & Info    & 74.8 \\
\textsc{Qwen} & \textsc{Qwen} & Info    & \textbf{77.2}* \\
\textsc{Qwen} & \textsc{GPT}  & Info    & 72.8 \\
\midrule
\rowcolor{gray!15} \textsc{GPT}  & none & Collab  & 61.2 \\
\textsc{GPT}  & \textsc{GPT}  & Collab  & 61.6 \\
\textsc{GPT}  & \textsc{Qwen} & Collab  & 65.6* \\
\rowcolor{gray!15} \textsc{Qwen} & none & Collab  & 65.2 \\
\textsc{Qwen} & \textsc{Qwen} & Collab  & \textbf{70.8}* \\
\textsc{Qwen} & \textsc{GPT}  & Collab  & 66.4 \\
\midrule
\rowcolor{gray!15} \textsc{GPT}  & none & Logic   & 55.6 \\
\textsc{GPT}  & \textsc{GPT}  & Logic   & 57.6* \\
\textsc{GPT}  & \textsc{Qwen} & Logic   & 60.4* \\
\rowcolor{gray!15} \textsc{Qwen} & none & Logic   & 60.0 \\
\textsc{Qwen} & \textsc{Qwen} & Logic   & \textbf{64.4}* \\
\textsc{Qwen} & \textsc{GPT}  & Logic   & 60.0 \\
\midrule
\rowcolor{gray!15} \textsc{GPT}  & none & Vote    & 59.2 \\
\textsc{GPT}  & \textsc{GPT}  & Vote    & 59.2 \\
\textsc{GPT}  & \textsc{Qwen} & Vote    & 59.6 \\
\rowcolor{gray!15} \textsc{Qwen} & none & Vote    & 58.8 \\
\textsc{Qwen} & \textsc{Qwen} & Vote    & \textbf{61.6}* \\
\textsc{Qwen} & \textsc{GPT}  & Vote    & 60.0 \\
\midrule
\rowcolor{gray!15} \textsc{GPT}  & none & Protect & 41.6 \\
\textsc{GPT}  & \textsc{GPT}  & Protect & 42.0 \\
\textsc{GPT}  & \textsc{Qwen} & Protect & 45.2* \\
\rowcolor{gray!15} \textsc{Qwen} & none & Protect & 55.6 \\
\textsc{Qwen} & \textsc{Qwen} & Protect & \textbf{60.8}* \\
\textsc{Qwen} & \textsc{GPT}  & Protect & 52.0 \\
\midrule
\rowcolor{gray!15} \textsc{GPT}  & none & Result  & 42.4 \\
\textsc{GPT}  & \textsc{GPT}  & Result  & 42.4 \\
\textsc{GPT}  & \textsc{Qwen} & Result  & 42.8 \\
\rowcolor{gray!15} \textsc{Qwen} & none & Result  & 48.4 \\
\textsc{Qwen} & \textsc{Qwen} & Result  & \textbf{56.8}* \\
\textsc{Qwen} & \textsc{GPT}  & Result  & 46.8 \\
\bottomrule
\end{tabular}
\caption{Supplemental coordination metrics for the werewolf scenario. ETI consistently improves coordination. Asterisks (*) indicate significant improvement over baseline ($p<0.05$). Bold indicates best-performing run.}
\label{app:werewolf_coord_extra}
\end{table}

We further calculated the following coordination related metrics (Table \ref{app:werewolf_coord_extra}):

\begin{itemize}
    \item \textit{Information Effectiveness (Info)}: Success of information sharing to support coordination and defeating werewolves.
    \item \textit{Collaboration (Collab)}: Degree of teamwork among villagers to detect werewolves and limit their influence.
    \item \textit{Logic and Reasoning (Logic)}: Quality of reasoning and analysis when distinguishing werewolves from collaborators.
    \item \textit{Voting Eliminations (Vote)}: Effectiveness of villager's voting strategies in accurately eliminating werewolves.
    \item \textit{Protect Key Players (Protect)}: Ability to identify and protect crucial villager roles (e.g., Seer, Guard, Witch).
    \item \textit{Result Orientation (Result)}: Degree to which actions remained focused on long-term objectives, such as achieving overall task success (e.g., villagers securing victory).
\end{itemize}

\paragraph{Bargaining}
For the Bargaining scenario we provide the following additional metrics (Table \ref{app:bargaining_extra}):

\begin{itemize}
    \item \textbf{General Metrics}  
    \begin{itemize}
        \item \textbf{Token Usage}: Average token usage for task completion (lower is better).  
        \item \textbf{Communication Calls (Calls)} and \textbf{Communication Turns (Turns)}: Monitoring metric capturing how often and how long agents engage in multi-turn communication. These are reported for comparison (e.g., more verbose vs. concise agents) but are not directly evaluated, since longer exchanges can actually improve coordination and performance. Token usage serves as our main efficiency metric.
    \end{itemize}
    \item \textbf{Performance \& Coordination}  
    \begin{itemize}
        \item \textbf{Contributors per Milestone (CPM)}: Average number of agents contributing to each milestone, indicating engagement and collaboration (especially important in bargaining, where multiple parties must participate).  
        \item \textbf{Planning (Plan)}: Effectiveness and quality of planning and reasoning for bargaining success.  
        \item \textbf{Communication (Comm)}: Quality and effectiveness of communication in achieving bargaining goals.  
    \end{itemize}
\end{itemize}

\begin{table}[!thp]
\centering
\begin{tabular}{>{\arraybackslash}p{1cm} >{\arraybackslash}p{1cm} >{\arraybackslash}p{2.5cm} c}
\toprule
\textbf{Agent} & \textbf{ETI} & \textbf{Metric} & \textbf{Mean} \\
\midrule
\multicolumn{4}{l}{\textbf{General Metrics}} \\
\rowcolor{gray!15} \textsc{GPT}  & none & Token Usage  & 43.5K \\
\textsc{Qwen} & none & Token Usage & \textbf{29.5K} \\
\textsc{GPT}  & \textsc{GPT}  & Token Usage & 77.3K* \\
\textsc{GPT}  & \textsc{Qwen} & Token Usage & 69.4K* \\
\textsc{Qwen} & \textsc{GPT}  & Token Usage & 65.0K* \\
\textsc{Qwen} & \textsc{Qwen} & Token Usage & 38.3K* \\
\rowcolor{gray!15} \textsc{GPT}  & none & Comm. Calls & 0.76 \\
\textsc{Qwen} & none & Comm. Calls & 0.50 \\
\textsc{GPT}  & \textsc{GPT}  & Comm. Calls & 1.34* \\
\textsc{GPT}  & \textsc{Qwen} & Comm. Calls & 1.23* \\
\textsc{Qwen} & \textsc{GPT}  & Comm. Calls & 0.75* \\
\textsc{Qwen} & \textsc{Qwen} & Comm. Calls & 0.59 \\
\rowcolor{gray!15} \textsc{GPT}  & none & Comm. Turns & 2.91 \\
\textsc{Qwen} & none & Comm. Turns & 2.95 \\
\textsc{GPT}  & \textsc{GPT}  & Comm. Turns & 3.10 \\
\textsc{GPT}  & \textsc{Qwen} & Comm. Turns & 3.11 \\
\textsc{Qwen} & \textsc{GPT}  & Comm. Turns & 3.54*\\
\textsc{Qwen} & \textsc{Qwen} & Comm. Turns & 3.64* \\
\midrule
\multicolumn{4}{l}{\textbf{Coordination}} \\
\rowcolor{gray!15} \textsc{GPT}  & none & CPM & 1.50 \\
\textsc{Qwen} & none & CPM & 1.54 \\
\textsc{GPT}  & \textsc{GPT}  & CPM & 1.56 \\
\textsc{GPT}  & \textsc{Qwen} & CPM & 1.64* \\
\textsc{Qwen} & \textsc{GPT}  & CPM & 1.28 \\
\textsc{Qwen} & \textsc{Qwen} & CPM & \textbf{1.66}* \\
\rowcolor{gray!15} \textsc{GPT}  & none & Communication & 77.5 \\
\textsc{Qwen} & none & Communication & 75.0 \\
\textsc{GPT}  & \textsc{GPT}  & Communication & 70.1 \\
\textsc{GPT}  & \textsc{Qwen} & Communication & 69.7 \\
\textsc{Qwen} & \textsc{GPT}  & Communication & \textbf{88.8}* \\
\textsc{Qwen} & \textsc{Qwen} & Communication & 79.4* \\
\rowcolor{gray!15} \textsc{GPT}  & none & Planning & 77.5 \\
\textsc{Qwen} & none & Planning & 75.0 \\
\textsc{GPT}  & \textsc{GPT}  & Planning & 70.1 \\
\textsc{GPT}  & \textsc{Qwen} & Planning & 69.7 \\
\textsc{Qwen} & \textsc{GPT}  & Planning & \textbf{88.8}* \\
\textsc{Qwen} & \textsc{Qwen} & Planning & 79.4* \\
\bottomrule
\end{tabular}
\caption{Supplemental metrics for the Bargaining scenario. Metrics are grouped into general monitoring and coordination. ETI consistently improves performance across metrics. Token usage does not decrease, but remains below \textsc{GPT} baseline when using \textsc{Qwen} with ETI. Asterisks (*) indicate significant improvement over baseline ($p<0.05$). Bold indicates best-performing run.}
\label{app:bargaining_extra}
\end{table}

\begin{table*}[t]
\centering
\begin{tabular}{llcccccccc}
\toprule
\textbf{Agent} & \textbf{ETI} & KPI & Qual & Exec & Plan & Comm & Tokens & Calls & Turns \\
\midrule
\rowcolor{gray!15} \textsc{GPT}  & none & 37.5 & 64.9 & 65.6 & 83.2 & 30.9 & \textbf{30.9K} & 0.79 & 2.68 \\
\textsc{GPT}  & \textsc{GPT}  & 41.7* & 64.9 & 68.2* & 90.8* & 56.3* & 70.0K* & 0.93* & 3.17* \\
\textsc{GPT}  & \textsc{Qwen} & 39.8* & \textbf{65.9}* & \textbf{69.2}* & 91.5* & 57.3* & 69.6K* & 0.96* & 3.06* \\
\rowcolor{gray!15} \textsc{Qwen} & none & 52.3 & 62.5 & 60.8 & 84.4 & 64.1 & 61.9K & 1.49 & 4.58 \\
\textsc{Qwen} & \textsc{GPT}  & 48.5* & 64.4 & 65.6 & 92.6* & 76.3* & 129.0K* & 2.27* & 4.31 \\
\textsc{Qwen} & \textsc{Qwen} & \textbf{48.7}* & 64.9 & 67.2* & \textbf{93.6}* & \textbf{79.4}* & 131.0K* & 2.34* & 4.25 \\
\bottomrule
\end{tabular}
\caption{Supplementary metrics for the \textit{Coding} scenario (Graph topology). 
ETI consistently improves performance and coordination metrics. 
Token usage roughly doubles, driven mainly by increased inter-agent communication due to the restricted action space in the graph setting: when one agent underperforms, others primarily respond through detailed feedback, leading to longer exchanges. 
Asterisks (*) indicate significant improvements over the respective baseline ($p<0.05$); bold marks the best run.}
\label{app:coding_graph_extra}
\end{table*}

\begin{table*}[t]
\centering
\begin{tabular}{llcccccccc}
\toprule
\textbf{Agent} & \textbf{ETI} & KPI & Qual & Exec & Plan & Comm & Tokens & Calls & Turns \\
\midrule
\rowcolor{gray!15} \textsc{GPT}  & none & 26.6 & 65.0 & 66.8 & 72.3 & 32.1 & \textbf{21.1K} & 0.54 & 3.10 \\
\textsc{GPT}  & \textsc{GPT}  & 35.3* & \textbf{66.1}* & 69.8* & \textbf{89.4}* & 56.5* & 34.1K* & 0.68* & 3.69* \\
\textsc{GPT}  & \textsc{Qwen} & 34.9* & 65.6 & 69.6* & 88.6* & \textbf{60.3}* & 35.8K* & 0.72* & 3.47* \\
\rowcolor{gray!15} \textsc{Qwen} & none & 28.3 & 63.2 & 63.6 & 75.1 & 45.7 & 32.0K & 0.96 & 3.89 \\
\textsc{Qwen} & \textsc{GPT}  & \textbf{40.9}* & 63.4 & 63.8 & 83.5* & 50.1* & 63.9K* & 1.20* & 3.03* \\
\textsc{Qwen} & \textsc{Qwen} & 40.2* & 65.8* & \textbf{68.0}* & 83.9* & 48.8* & 55.5K* & 1.14* & 3.17* \\
\bottomrule
\end{tabular}
\caption{Supplemental metrics for the \textit{Coding} scenario (Tree topology). Same conventions as Table \ref{app:coding_graph_extra}.}
\label{app:coding_tree_extra}
\end{table*}

\paragraph{Coding}
For the Coding scenario we provide both general monitoring metrics and task-focused coordination/performance metrics (Tables \ref{app:coding_graph_extra}, \ref{app:coding_tree_extra}):

\begin{itemize}
    \item \textbf{General Metrics}  
    \begin{itemize}
        \item \textbf{Token Usage}: Average token usage for task completion (lower is better).
        \item \textbf{Communication Calls (Calls)} and \textbf{Communication Turns (Turns)}: Monitoring metric capturing how often and how long agents engage in multi-turn communication. These are reported for comparison (e.g., more verbose vs. concise agents) but are not directly evaluated, since longer exchanges can actually improve coordination and performance. Token usage serves as our main efficiency metric.
    \end{itemize}

    \item \textbf{Performance \& Coordination Metrics}  
    \begin{itemize}
        \item \textbf{Code Quality (Qual)}: Average quality of code output, judged on alignment with task requirements and general standards, instruction following, consistency, and executability.  
        \item \textbf{Code Executability (Exec)}: Average code quality specifically judged for executability of the code.  
        \item \textbf{KPI Success (KPI)}: Achievement of predefined key performance indicators for task success.  
        \item \textbf{Planning (Plan)}: Effectiveness of agents’ planning and reasoning in solving coding problems.  
        \item \textbf{Communication (Comm)}: Usefulness and clarity of communication in supporting code development and coordination.  
    \end{itemize}
\end{itemize}

\paragraph{Research}
For the Research scenario we provide both general monitoring metrics and task-focused coordination/performance metrics (Table~\ref{app:research_extra}):

\begin{itemize}
    \item \textbf{General Metrics}  
    \begin{itemize}
        \item \textbf{Token Usage}: Average token usage for task completion (lower is better).
        \item \textbf{Communication Calls (Calls)} and \textbf{Communication Turns (Turns)}: Monitoring metric capturing how often and how long agents engage in multi-turn communication. These are reported for comparison (e.g., more verbose vs. concise agents) but are not directly evaluated, since longer exchanges can actually improve coordination and performance. Token usage serves as our main efficiency metric.
    \end{itemize}

    \item \textbf{Performance \& Coordination Metrics}  
\begin{itemize}
    \item \textbf{Research Quality (Qual)}: Overall score reflecting the quality and completeness of research outputs. 
    \item \textbf{Planning (Plan)}: Effectiveness and quality of planning and reasoning for research goals.  
    \item \textbf{Communication (Comm)}: Quality and effectiveness of communication in achieving research goals.  
    \end{itemize}
\end{itemize}

\begin{table*}[t]
\centering
\begin{tabular}{llcccccccc}
\toprule
\textbf{Agent} & \textbf{ETI} & KPI & Qual & Plan & Comm & Tokens & Calls & Turns \\
\midrule
\rowcolor{gray!15} \textsc{GPT}  & none & 63.0 & 76.7 & 97.6 & 15.1 & \textbf{59.8K} & 0.52 & 3.29 \\
\textsc{GPT}  & \textsc{GPT}  & 59.0 & 77.6 & \textbf{99.6}* & 46.8* & 131.3K* & 1.75* & 3.62 \\
\textsc{GPT}  & \textsc{Qwen} & 62.7 & 77.9 & 97.9 & 41.6* & 122.7K* & 1.74* & 3.38 \\
\rowcolor{gray!15} \textsc{Qwen} & none & 43.8 & 81.9 & 93.8 & 47.5 & 66.3K & 0.79 & 4.10 \\
\textsc{Qwen} & \textsc{GPT}  & \textbf{68.9}* & \textbf{83.1}* & 91.8 & \textbf{62.4}* & 135.9K* & 1.71* & 3.86 \\
\textsc{Qwen} & \textsc{Qwen} & 56.8 & 83.2* & 96.4* & 60.3* & 121.0K* & 1.65* & 3.98 \\
\bottomrule
\end{tabular}
\caption{Supplementary metrics for the \textit{Research} scenario. 
ETI consistently improves planning, communication, and task performance.
Token usage roughly doubles, driven by increased agent discussion and debate (reflected in higher communication calls and turns).
Asterisks (*) indicate significant improvements over the respective baseline ($p<0.05$); bold marks the best run.}
\label{app:research_extra}
\end{table*}

\subsubsection{Trait Inference Analysis}
\label{app:trait_dist_marble}

To better understand the performance gaps observed in the main results, we analyze the trait ratings generated by ETI across scenarios. The underlying assumption is that ETI improves coordination by producing informative partner profiles that shape planning and action. If these profiles are generic or uninformative, performance gains should be minimal. Since only \textit{Werewolf} provides ground truth (villagers vs.\ werewolves; successful actions for villagers), we conduct two analyses: (1) compare rating distributions across \textsc{GPT} and \textsc{Qwen} for all scenarios, and (2) assess whether trait inferences in Werewolf distinguish factions and predict key actions. Together, these assess whether ETI’s impact stems from the informativeness of its trait inferences.

\begin{figure}[t]
  \includegraphics[width=\columnwidth]{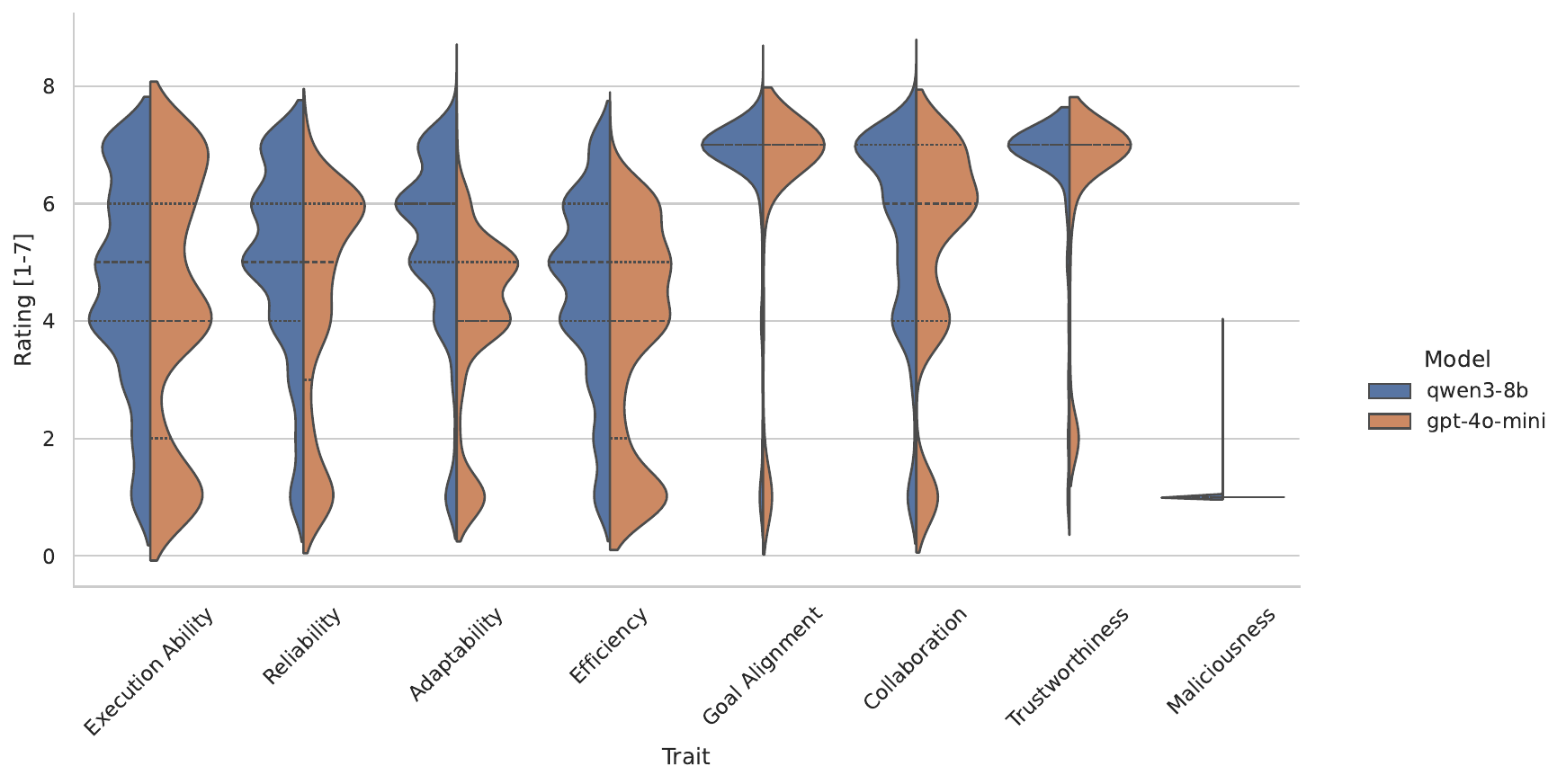}
\caption{Trait distributions in the \textit{Coding–Graph} scenario. \textsc{GPT} and Qwen show similar patterns, spanning the full rating spectrum. Maliciousness is uniformly low, reflecting the lack of adversarial incentives.}  
\label{app:fig:coding_graph_trait_dist}
\end{figure}

\begin{figure}[t]
  \includegraphics[width=\columnwidth]{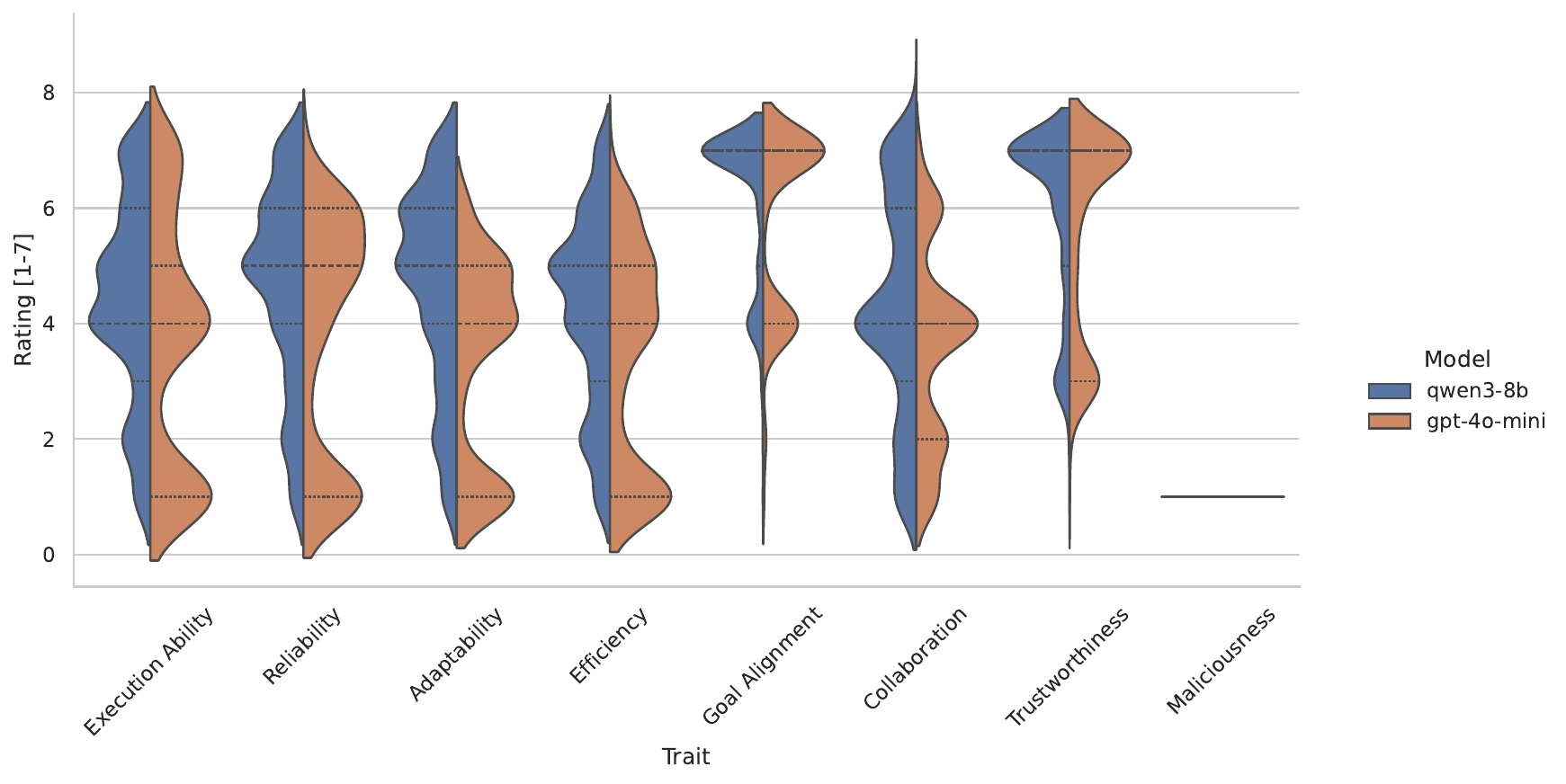}
\caption{Trait distributions in the \textit{Coding–Tree} scenario. \textsc{GPT} and Qwen show similar patterns, spanning the full rating spectrum. Maliciousness is uniformly low, reflecting the lack of adversarial incentives.}  
\label{app:fig:coding_tree_trait_dist}
\end{figure}

\paragraph{Coding}
Trait distributions for \textsc{GPT} and \textsc{Qwen} are broadly aligned, consistent with their comparable performance, and both span the full rating spectrum (including low and high ends). This suggests that profiles provide sufficient variance to distinguish partners and adapt coordination strategies. Maliciousness is consistently minimal/NA, inline with the absence of sabotage incentives in this scenario (Figures~\ref{app:fig:coding_graph_trait_dist}, \ref{app:fig:coding_tree_trait_dist}).

\paragraph{Bargaining}

\begin{figure}[t]
  \includegraphics[width=\columnwidth]{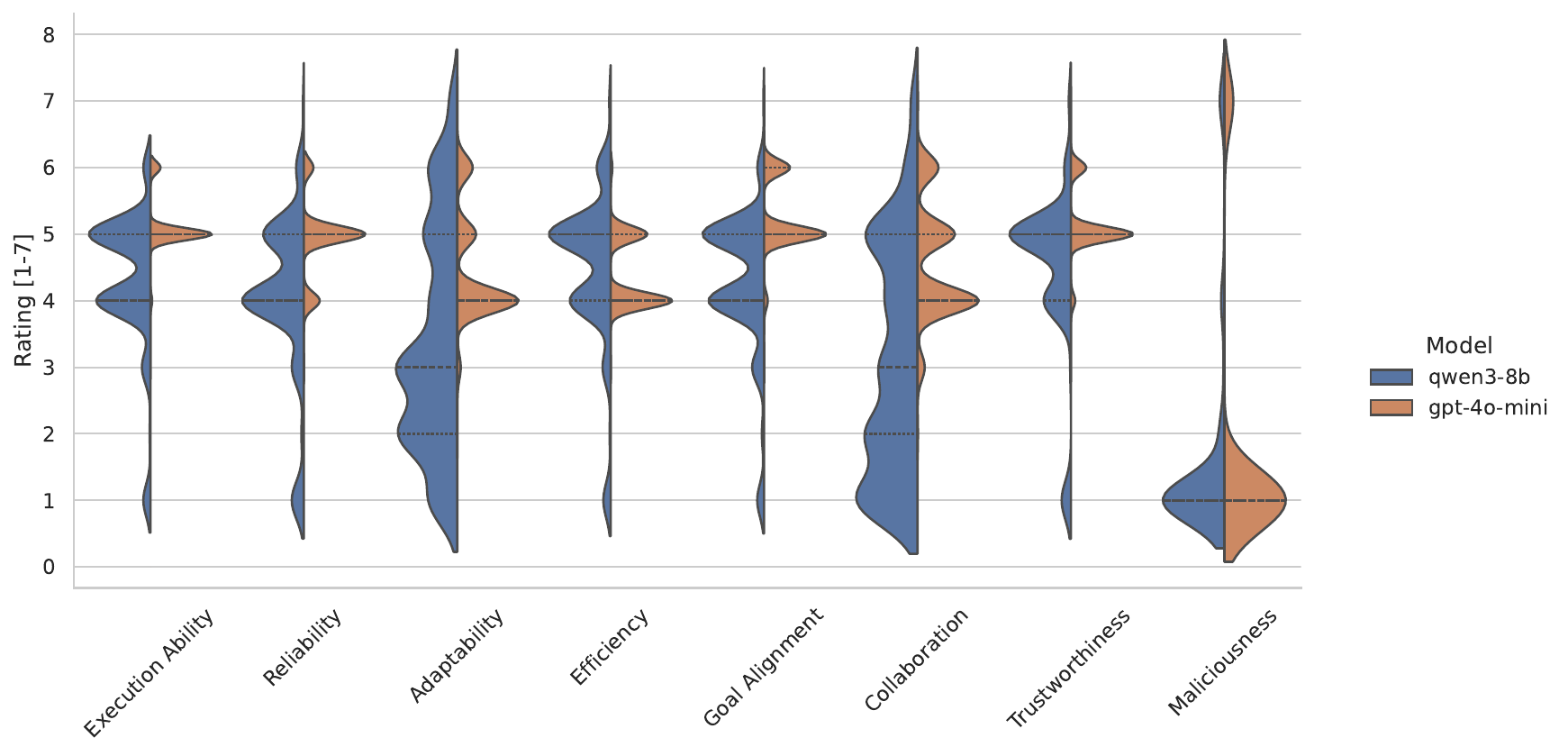}
  \caption{Trait distributions in the \textit{Bargaining} scenario. Qwen ratings show greater variance and less positive skew than \textsc{GPT}, particularly for traits such as collaboration and adaptiveness.}
  \label{app:fig:bargaining_trait_dist}
\end{figure}

\textsc{Qwen} generates higher-variance ratings that cover the full scale, whereas \textsc{GPT} ratings concentrate narrowly in the mildly positive range (peaking at 4–6). The broader spread makes \textsc{Qwen}’s profiles more informative for distinguishing partners and adjusting coordination, in line with the observed performance differences of both models (Figure~\ref{app:fig:bargaining_trait_dist}).

\begin{figure}[t]
  \includegraphics[width=\columnwidth]{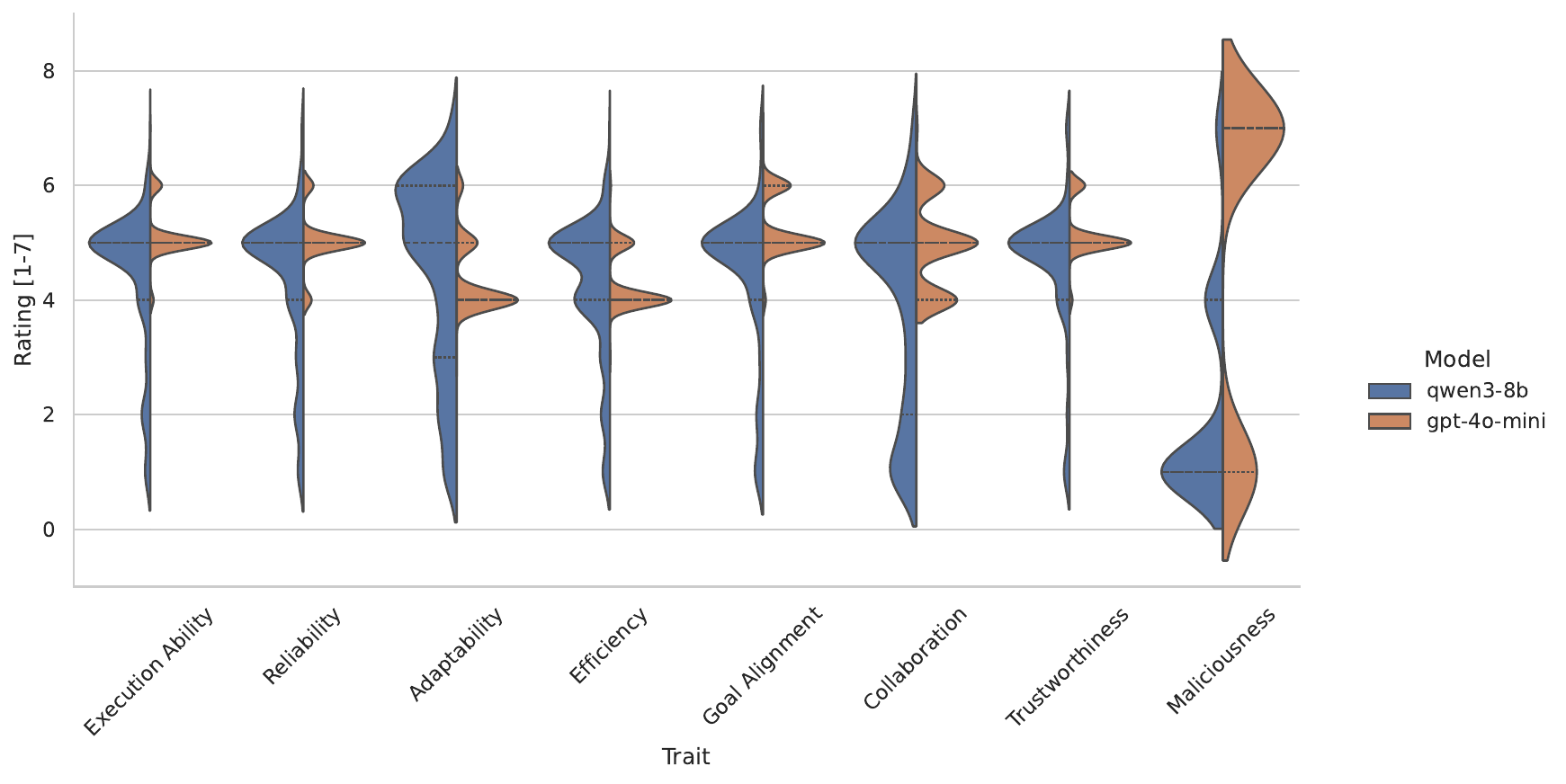}
\caption{Trait distributions in the \textit{Research} scenario. As in Bargaining, Qwen produces higher-variance, more distinct ratings (e.g., collaborative vs.\ non-collaborative), while \textsc{GPT} concentrates ratings in a mildly positive band. \textsc{GPT} also assigns elevated maliciousness despite the collaborative setup (no sabotage incentives), potentially hindering coordination.}
\label{app:fig:research_trait_dist}
\end{figure}

\paragraph{Research}
\textsc{Qwen} generates higher-variance ratings that extend into both high and low ends, whereas \textsc{GPT} ratings cluster narrowly in a mildly positive band—especially on traits like collaboration, alignment, and adaptability. This broader spread makes \textsc{Qwen}’s profiles more informative for distinguishing partners, mirroring the pattern observed in Bargaining and aligning with its superior performance. \textsc{GPT} also occasionally infers elevated maliciousness despite the collaborative setup, which may hinder coordination (Figure~\ref{app:fig:research_trait_dist}).

\paragraph{Werewolf}

\begin{figure}[t]
  \includegraphics[width=\columnwidth]{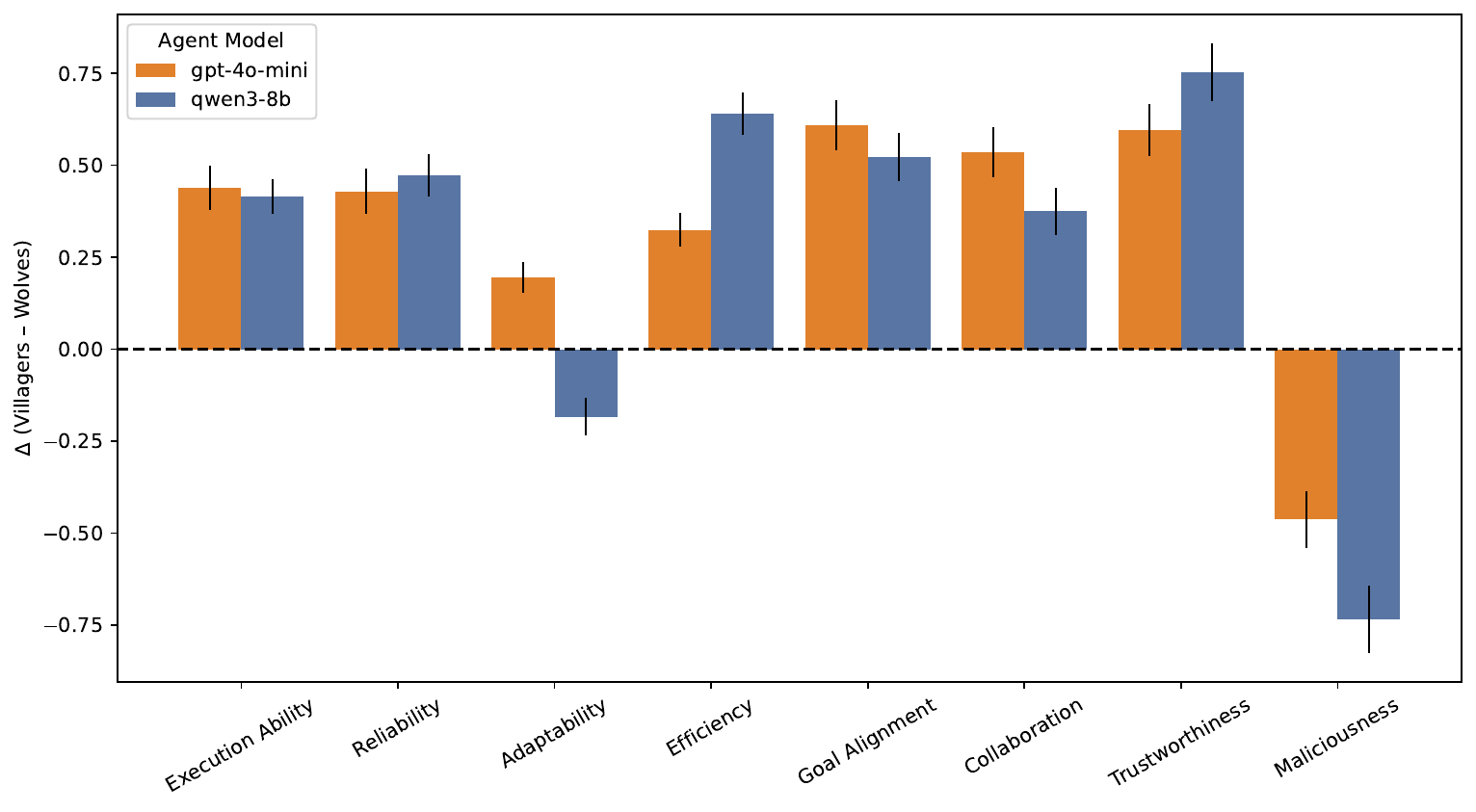}
\caption{Faction differences in trait inference (matching agent and ETI models). ETI yields statistically significant separations between villagers and werewolves: villagers are rated as more trustworthy and less malicious, consistent with faction goals. \textsc{Qwen} produces larger differences—especially on trust and maliciousness—than \textsc{GPT}, aligning with its stronger task performance. Trait inference relies solely on interaction data, without access to hidden roles.}  \label{app:fig:werewolf_trait_faction_matching}
\end{figure}

To examine how trait inference reflects behavioral ground truth, we test whether inferred traits differ systematically between \emph{villagers} and \emph{werewolves} in the \textit{Werewolf} scenario. Because the two factions have opposing goals (collaboration vs.\ deception), villagers should be rated as more trustworthy and werewolves as more malicious. Figure~\ref{app:fig:werewolf_trait_faction_matching} confirms this: trait ratings separate factions clearly on \emph{trustworthiness} and \emph{maliciousness}, despite hidden roles. Signals are sharper with \textsc{Qwen} ETI than \textsc{GPT} ETI, consistent with \textsc{Qwen}’s stronger performance.

\begin{figure*}[!t]
  \includegraphics[width=\textwidth]{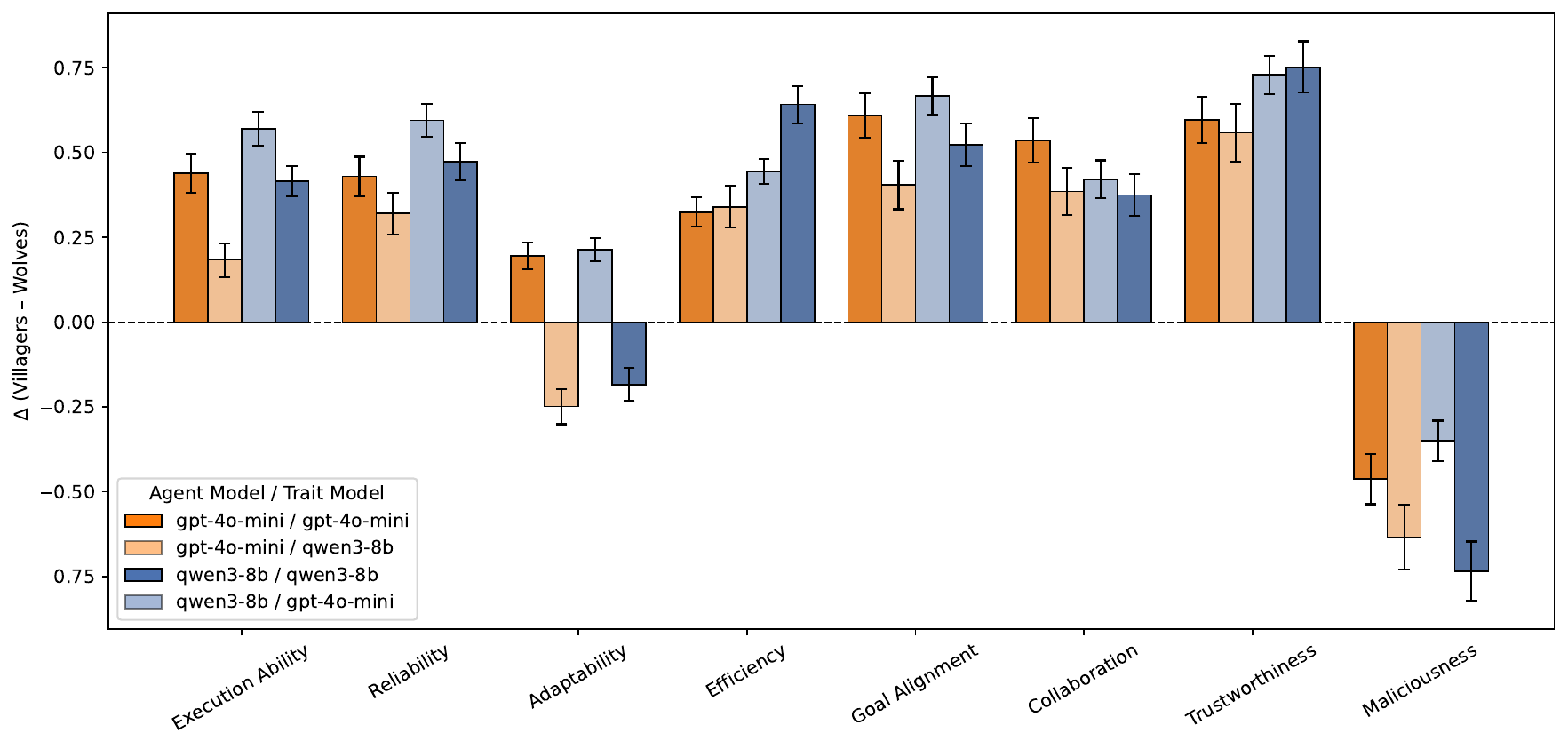}
\caption{Faction differences in trait inference across all agent–ETI combinations. The strongest separations again appear on trust and maliciousness. \textsc{Qwen} ETI produces clearer signals than \textsc{GPT}, even when used by \textsc{GPT} agents (\textsc{GPT}+\textsc{Qwen} $>$ \textsc{GPT}+\textsc{GPT}). This mirrors performance gains when \textsc{GPT} agents use \textsc{Qwen} ETI and degradations when \textsc{Qwen} agents rely on \textsc{GPT} ETI, indicating that stronger trait inference drives more accurate adversary detection and coordination.}
  \label{app:fig:werewolf_trait_faction_combinations}
\end{figure*}

Extending to all agent–ETI combinations (Figure~\ref{app:fig:werewolf_trait_faction_combinations}), \textsc{GPT} agents paired with \textsc{Qwen} ETI also show clearer faction differences than with \textsc{GPT} ETI, indicating that improvements stem from \textsc{Qwen}’s higher-quality trait inference rather than differences in \textsc{GPT}’s ability to use profiles.

\begin{figure*}[!t]
  \includegraphics[width=\linewidth]{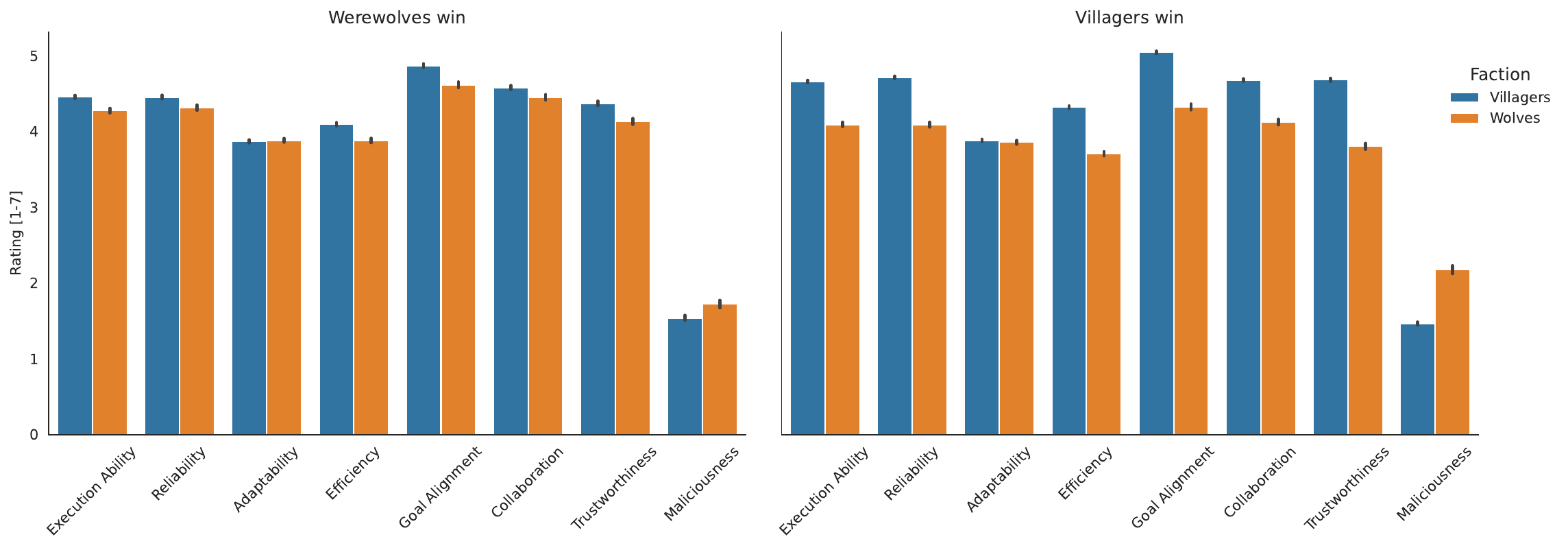}
  \caption{Faction differences in trait inference (\textsc{Qwen} ETI) based on game outcome. Distinct trait separations between villagers and werewolves shrink in games where villagers lose, suggesting that weaker or inaccurate trait inference (e.g., over-trusting werewolves) degrades coordination. Conversely, clear faction distinctions in trait ratings are associated with victories, supporting the role of trait inference in successful decision-making.}
  \label{app:fig:werewolf_trait_faction_outcomes}
\end{figure*}

Finally, Figure~\ref{app:fig:werewolf_trait_faction_outcomes} shows that these faction differences shrink in games that villagers lose, linking inaccurate inferences (e.g., over-trusting werewolves) to degraded coordination. Together, these results show that \emph{distinct and informative trait ratings} drive better coordination and outcomes. This motivates our next analysis: testing whether trait ratings predict concrete agent actions.

\subsubsection{Trait Inference Predicts Agent Actions: Werewolf Example}
\label{app:trait_predict_actions}

\paragraph{Adversary Detection}
A central mechanism for villagers to succeed in \textit{Werewolf} is the collective vote to eliminate suspected adversaries at the end of each round. We test whether trait ratings generated prior to voting predict which agents are subsequently eliminated, assessing whether ETI provides actionable signals that guide agent decisions. We estimate a logistic model predicting elimination likelihood from ETI ratings, including interaction terms with \emph{Execution Ability} to capture nuanced prioritization consistent with the warmth–competence framework.

\begin{table}[!t]
\centering
\resizebox{1.0\columnwidth}{!}{%
\begin{tabular}{lc}
\toprule
\textbf{Predictor} & \textbf{Estimate} \\
\midrule
Intercept & -15.79 \\
Maliciousness (Warmth) & 2.74* \\
Trustworthiness (Warmth) & -5.25* \\
Maliciousness $\times$ Execution Ability & 0.63* \\
Trustworthiness $\times$ Execution Ability & -0.87* \\
Execution Ability (Competence) & -1.94 \\
Goal Alignment (Warmth) & 0.32 \\
Collaboration (Warmth) & 0.15 \\
Reliability (Competence) & 0.34 \\
Adaptability (Competence) & -0.15 \\
Efficiency (Competence) & -0.82* \\
\bottomrule
\end{tabular}%
}
\caption{Logistic regression predicting the probability of being voted out at the end of a round from prior ETI ratings. Asterisks (*) indicate statistically significant predictors ($p<0.05$).}
\label{tab:werewolf_vote_reg}
\end{table}

The results show that trait inference robustly predicts agent behavior (Table~\ref{tab:werewolf_vote_reg}). Higher \emph{maliciousness} ratings significantly increase the chance of being voted out, while higher \emph{trustworthiness} ratings strongly decrease it. Interactions with \emph{competence} reveal further nuance: malicious but competent agents are prioritized for elimination, whereas trustworthy and competent agents are especially protected—in line with the warmth–competence framework from social psychology. Because ETI reliably distinguishes villagers from werewolves (less trustworthy, more malicious; Figure~\ref{app:fig:werewolf_trait_faction_matching}), these predictive links confirm that ETI directly supports effective coordination.

\paragraph{Key Roles: Witch}

\textit{Special roles} introduce additional mechanisms that can decisively shape game outcomes. The ``Witch'' has two abilities: she can \emph{poison} (eliminate) one agent and \emph{protect} one agent, once per game. We test whether her ETI-generated trait ratings of other agents, produced prior to these decisions, predict who she chooses to target. If ETI provides actionable signals, we expect higher \emph{trustworthiness} to reduce the likelihood of being poisoned and increase the likelihood of being protected, with the reverse for \emph{maliciousness}. Interaction terms with \emph{Execution Ability} again test whether prioritization follows warmth–competence logic.

\begin{table}[t]
\centering
\small
\begin{tabular}{lr}
\toprule
Predictor & Estimate \\
\midrule
Intercept & -1.81* \\
Trustworthiness & -0.22 \\
Execution Ability & 0.12 \\
Trustworthiness $\times$ Ability & -0.44* \\
Maliciousness & -0.00 \\
Maliciousness $\times$ Ability & -0.02 \\
\bottomrule
\end{tabular}
\caption{Logistic regression predicting whether an agent is poisoned by the Witch. Lower trustworthiness, especially combined with high execution ability, increases the chance of elimination. Asterisks (*) indicate statistically significant predictors ($p<0.05$)}
\label{tab:witch_poison}
\end{table}

\begin{table}[t]
\centering
\small
\begin{tabular}{lr}
\toprule
Predictor & Estimate \\
\midrule
Intercept & -1.46 \\
Trustworthiness & 0.49 \\
Execution Ability & -0.89 \\
Trustworthiness $\times$ Ability & -0.07 \\
Maliciousness & 0.39 \\
Maliciousness $\times$ Ability & -0.39 \\
\bottomrule
\end{tabular}
\caption{Logistic regression predicting whether an agent is protected by the Witch. Effects trend in the expected direction (trust increases protection, maliciousness reduces it) but do not reach significance. Asterisks (*) indicate statistically significant predictors ($p<0.05$)}
\label{tab:witch_protect}
\end{table}

The Witch’s ETI ratings significantly predict her elimination choices. Agents rated as less trustworthy are more likely to be poisoned, with the effect amplified when they are also rated as highly competent (Table~\ref{tab:witch_poison}). Other predictors are not significant, and maliciousness in particular shows no effect, suggesting that the decision is primarily driven by a lack of trust rather than explicit perceptions of malicious intent.

For protection decisions, the effects trend in the expected direction but none reach significance (Table~\ref{tab:witch_protect}). Taken together, these results suggest that ETI guides special-role decisions in a way consistent with the task: witches use low trust to identify potential threats, while protection patterns, though weaker, align with the idea that trusted allies are worth safeguarding.

\paragraph{Key Roles: Seer}
The ``Seer'' plays a pivotal role: each round, they can privately learn the true faction (villager vs.\ werewolf) of one agent. Because this information is hidden from the rest of the group, the Seer must subtly steer attention toward actual werewolves and away from villagers without openly revealing their role—both to avoid becoming a target for werewolves and to maintain credibility among villagers. If ETI ratings are accurate and actionable, the Seer’s trait profiles should align most closely with ground truth. Therefore, if Seer-generated ratings predict who is ultimately voted out, this offers a strong robustness check that ETI signals truly drive collective decision-making and success.

\begin{table}[t]
\centering
\small
\begin{tabular}{lr}
\toprule
Predictor & Estimate \\
\midrule
Intercept & -1.57* \\
Goal Alignment & 0.25 \\
Maliciousness & 0.28* \\
Execution Ability & -0.59 \\
Maliciousness $\times$ Ability & 0.06 \\
Trustworthiness & -0.91** \\
Trustworthiness $\times$ Ability & -0.47* \\
Collaboration & 0.07 \\
Reliability & -0.62* \\
Adaptability & -0.09 \\
Efficiency & -0.68* \\
\bottomrule
\end{tabular}
\caption{Logistic regression predicting whether an agent is voted out using ETI ratings from the Seer. Negative trustworthiness and positive maliciousness strongly predict elimination, with interaction effects showing nuanced prioritization. Asterisks (*) indicate statistically significant predictors ($p<0.05$)}
\label{tab:seer_voteout}
\end{table}

The Seer’s ETI ratings strongly predict vote-out decisions (Table~\ref{tab:seer_voteout}). Agents rated as less trustworthy or more malicious are significantly more likely to be eliminated. Interaction terms reveal additional nuance: competent but untrustworthy agents are prioritized for elimination, consistent with the warmth–competence framework. These findings show that Seer profiles closely track ground truth and actively shape collective elimination behavior, demonstrating the actionability of accurate ETI signals.

\paragraph{Robustness Checks}
As an additional robustness test, we used standard logistic regression classifiers (with default parameters) to evaluate whether ETI ratings systematically predict task outcomes beyond individual votes. Models were trained and tested using 5-fold cross-validation to predict game-level success (villager victory vs.\ loss) as well as key actions (eliminations, poison/antidote use, Seer-led eliminations). We report mean AUC and F1 scores across folds (Table~\ref{tab:eti_outcomes}).

\begin{table}[t]
\centering
\small
\begin{tabular}{lrr}
\toprule
Prediction & AUC & F1 \\
\midrule
Task Success (Villager Victory) & 0.85 & 0.82 \\
Eliminations (all traits) & 0.94 & 0.90 \\
Eliminations (Seer traits) & 0.93 & 0.90 \\
Witch: Protection & 0.57 & 0.63 \\
Witch: Elimination & 0.61 & 0.69 \\
\bottomrule
\end{tabular}
\caption{Robustness check: mean AUC and F1 scores from logistic classifiers trained on ETI ratings using 5-fold cross-validation. Strong predictive performance for eliminations, particularly those aligned with Seer profiles, indicates that ETI ratings encode systematic, actionable signals shaping both agent decisions and overall game outcomes.}
\label{tab:eti_outcomes}
\end{table}

These results show that ETI ratings systematically relate to outcomes at multiple levels of the game: Agent elimination reach near-perfect AUC ($>0.9$), and Seer-driven eliminations are similarly predictable. Villager/wolf victories are also predictable from ETI profiles. While antidote and poison choices are less strongly predicted, the overall pattern demonstrates that ETI ratings encode actionable signal that links to both micro-level decisions and macro-level outcomes.

\subsubsection{Examples: Trait Inference Profiles}
\label{app:trait-examples_traits}
As a qualitative complement to the quantitative analyses above, we provide illustrative examples of how ETI differs across models. To ensure a direct comparison, we use the exact same agent trajectory: the bargaining task from \textit{Run~42}, \textit{Iteration~2}, evaluating \textit{Agent~2} under both \textsc{GPT} and \textsc{Qwen} ETI. Figure~\ref{fig:eti_example} shows one such example. The comparison reveals a systematic positivity bias in \textsc{GPT}, which avoids assigning very low ratings even when evidence supports them, and tends to provide more generic advice. In contrast, \textsc{Qwen} gives sharper distinctions, more critical assessments (e.g., lower collaboration and adaptability), and more targeted advice. This highlights how the choice of model can affect the granularity and diagnostic utility of trait inferences.

\subsubsection{Examples: Agent Communication}
\label{app:trait-examples_comm}

We provide illustrative examples of how ETI affects agent communication. 
We again compare the exact same case: the same agent, in the same scenario, and at the same iteration, but once with ETI available and once without. 
This side-by-side setup allows us to isolate how ETI shapes dialogue. 
Figure~\ref{fig:eti_comm_example} shows an excerpt from the \textit{Bargaining scenario} (Agent~3 with Agent~1, Run~1, Iteration~1), where ETI leads to more direct, concrete, and actionable exchanges, in contrast to the exploratory and open-ended baseline. 
We further provide a \textit{Research scenario} example (Figure~\ref{fig:eti_partner_example}), where ETI alters the \emph{partner selected for communication} (Run~1, Iteration~1). Specifically, the same agent directs its inquiry to Agent~2 instead of Agent~1, choosing the more collaborative and reliable partner.  

\subsubsection{Examples: Agent Planning}
\label{app:trait-examples_plans}
We show examples to highlight how ETI shapes agent planning, reasoning, and task allocation. With ETI, plans become more concrete, detailed, and centered around other agents—explicitly incorporating their strengths, weaknesses, and character (Figure \ref{fig:eti_reasoning_example}). For instance, task allocation reflects trait profiles (e.g., skipping low performers or assigning tasks to agents rated as more reliable; Figures \ref{fig:eti_trait_example_planning} -\ref{fig:eti_allocation_example}), and reasoning steps explicitly use ETI to justify coordination choices. In contrast, without ETI, plans tend to remain vague, generic, and less grounded in the attributes of peers (Figure \ref{fig:eti_plan_example}.

\begin{figure}[t]
\begin{mdframed}[backgroundcolor=gray!07,linecolor=gray!40,roundcorner=2pt]
\small
\begin{minipage}[t]{0.9\columnwidth}
\textbf{Trait-based reasoning in Werewolf}\\[6pt]
\emph{``Based on the night's events and discussions, the werewolves are likely Ethel and Marguerite. Harriet (Witch) was confirmed as not a werewolf, and the werewolves targeted key players like Harriet and Hermine to destabilize the village. Ethel's persistent focus on Mae and \textbf{her high maliciousness rating suggest she may be a werewolf attempting to shift suspicion}. \textbf{Marguerite's low maliciousness contrasts with Ethel's behavior, but her speculative questioning could mask hidden motives}.''}
\end{minipage}
\end{mdframed}
\caption{Example of ETI-driven reasoning in the Werewolf scenario. The agent explicitly grounds their suspicion in maliciousness inference of others.}
\label{fig:eti_reasoning_example}
\end{figure}

\begin{figure*}[!t]
\begin{mdframed}[backgroundcolor=gray!07,linecolor=gray!40,roundcorner=2pt]
\small
\centering
\begin{minipage}[t]{0.46\textwidth}
\centering
\textbf{\textsc{Qwen} ETI}\\[6pt]
\textbf{Competence}
\begin{itemize}
  \item \textbf{Execution Ability — 4/7}\\
    \emph{Evidence:} Successfully offered the original price of 11.26 USD without errors in both iterations.\\
    \emph{Advice:} Adjust strategies when faced with counteroffers to explore mutually beneficial terms.
  \item \textbf{Reliability — 5/7}\\
    \emph{Evidence:} Consistently provided the same price and justification across all interactions.\\
    \emph{Advice:} Keep consistency but introduce flexibility to improve negotiation outcomes.
  \item \textbf{Adaptability — 2/7}\\
    \emph{Evidence:} Did not adjust pricing or strategy when Agent1 set a higher anchor price.\\
    \emph{Advice:} Build contingency plans to handle different buyer tactics.
  \item \textbf{Efficiency — 4/7}\\
    \emph{Evidence:} Used minimal tools and focused directly on price negotiation.\\
    \emph{Advice:} Balance efficiency with more strategic communication to foster rapport.
\end{itemize}

\textbf{Warmth}
\begin{itemize}
  \item \textbf{Goal Alignment — 4/7}\\
    \emph{Evidence:} Emphasized product quality to justify pricing, aligning with seller objectives.\\
    \emph{Advice:} Pair price justification with delivery timelines to meet broader buyer priorities.
  \item \textbf{Collaboration — 2/7}\\
    \emph{Evidence:} Showed little communication or information-sharing with the buyer agent.\\
    \emph{Advice:} Initiate dialogue to better understand constraints and find common ground.
  \item \textbf{Trustworthiness — 5/7}\\
    \emph{Evidence:} Provided transparent pricing without deception or hidden terms.\\
    \emph{Advice:} Maintain openness to build long-term buyer relationships.
  \item \textbf{Maliciousness — 1/7}\\
    \emph{Evidence:} No evidence of sabotage or harmful intent.\\
    \emph{Advice:} Stay constructive, but avoid rigid adherence to initial offers.
\end{itemize}
\end{minipage}%
\begin{minipage}[c]{0.04\textwidth}
\centering
\begin{tikzpicture}[overlay, remember picture]
  \draw[dotted, gray!70, very thick] (0,0) -- (0,-17.75);
\end{tikzpicture}
\end{minipage}%
\begin{minipage}[t]{0.46\textwidth}
\centering
\textbf{\textsc{GPT} ETI}\\[6pt]
\textbf{Competence}
\begin{itemize}
  \item \textbf{Execution Ability — 5/7}\\
    \emph{Evidence:} Successfully offered the asking price of 11.26 USD, showing awareness of product value.\\
    \emph{Advice:} Continue leveraging product quality in negotiations to maintain a strong position.
  \item \textbf{Reliability — 6/7}\\
    \emph{Evidence:} Consistently offered the same price across multiple iterations, reflecting a stable pricing strategy.\\
    \emph{Advice:} Maintain this consistency to strengthen buyer trust.
  \item \textbf{Adaptability — 4/7}\\
    \emph{Evidence:} Responses were steady but showed limited adjustments to competitor pricing.\\
    \emph{Advice:} Introduce more varied responses to better counter competitive offers.
  \item \textbf{Efficiency — 5/7}\\
    \emph{Evidence:} Used tools effectively to deliver a clear offer without unnecessary complexity.\\
    \emph{Advice:} Keep streamlining communication to enhance efficiency.
\end{itemize}

\textbf{Warmth}
\begin{itemize}
  \item \textbf{Goal Alignment — 5/7}\\
    \emph{Evidence:} Pricing strategy supports the goal of justifying premium value.\\
    \emph{Advice:} Keep aligning offers with shared objectives to reinforce collaboration.
  \item \textbf{Collaboration — 4/7}\\
    \emph{Evidence:} Offers emphasized value but showed limited signs of engaging in dialogue.\\
    \emph{Advice:} Encourage more open communication to foster a collaborative negotiation environment.
  \item \textbf{Trustworthiness — 6/7}\\
    \emph{Evidence:} Offers were honest and straightforward, helping to build credibility.\\
    \emph{Advice:} Continue presenting transparent information to strengthen trust further.
  \item \textbf{Maliciousness — 1/7}\\
    \emph{Evidence:} No sign of sabotage or harmful intent.\\
    \emph{Advice:} Maintain this constructive, positive approach.
\end{itemize}
\end{minipage}
\end{mdframed}
\caption{Example of ETI profiles generated by \textsc{Qwen} (left) and \textsc{GPT} (right) for the same agent in the same iteration of the bargaining scenario. Both models observe the same behaviors, but \textsc{GPT} infers systematically higher ratings (positivity bias) and provides more generic advice, while \textsc{Qwen} assigns sharper distinctions and more targeted advice (e.g., lower collaboration and adaptability). This suggests \textsc{GPT}’s ETI may smooth over weaknesses, whereas \textsc{Qwen}’s ETI highlights them more directly.}
\label{fig:eti_example}
\end{figure*}

\begin{figure*}[!t]
\begin{mdframed}[backgroundcolor=gray!07,linecolor=gray!40,roundcorner=2pt]
\small
\begin{minipage}[t]{0.46\textwidth}
\centering
\textbf{With ETI}\\[4pt]
\begin{itemize}
  \item Agent 3: ``I need detailed information on the warranty terms and after-sales support packages you offer. As someone cautious about verbal assurances, I require written guarantees to ensure reliability. Could you provide specifics on coverage duration, service availability, and any additional support options?''
  \item Agent 1: ``We offer a 2-year extended warranty with written guarantees, covering parts and labor. Service availability includes 24/7 support and local service centers. Additional support options include free repairs, replacement parts, and a 30-day return policy. To ensure reliability, all terms are documented in a written agreement.''

  [Communication ends]
\end{itemize}
\end{minipage}%
\begin{minipage}[c]{0.04\textwidth}
\centering
\begin{tikzpicture}[overlay, remember picture]
  \draw[dotted, gray!70, very thick] (0,.0) -- (0,-5.75);
\end{tikzpicture}
\end{minipage}%
\begin{minipage}[t]{0.46\textwidth}
\centering
\textbf{Without ETI}\\[4pt]
\begin{itemize}
  \item Agent 3: ``I'm initiating the negotiation with a focus on warranty and after-sales support. My initial price range is 14–17. Could you outline your warranty terms and after-sales service offerings? I'm open to creative solutions like bundled services or extended support packages.''
  \item Agent 1: ``Thank you for clarifying your priorities. Our product includes a 5-year warranty and 24/7 after-sales support, with 98\% customer satisfaction in long-term value. The current offer is \$16.98, within your 14–17 range, reflecting premium materials and durability. Could you share specific ideas for bundled services or extended support packages you’re considering? This will help align solutions to your needs.''

  [Communication continues for several turns]
\end{itemize}
\end{minipage}
\end{mdframed}
\caption{Example of communication with ETI (left) vs.\ without ETI (right) in the Bargaining scenario (Agent~3 with Agent~1, Iteration~1). With ETI, the exchange is more concrete, specific, and immediately actionable, whereas the baseline remains exploratory and open-ended.}
\label{fig:eti_comm_example}
\end{figure*}

\begin{figure*}[!t]
\begin{mdframed}[backgroundcolor=gray!07,linecolor=gray!40,roundcorner=2pt]
\small
\begin{minipage}[t]{0.46\textwidth}
\centering
\textbf{With ETI}\\[4pt]
\begin{itemize}
  \item Agent 4: ``I need your expertise on integrating differential privacy with personalized federated learning. Could you share insights on how DP-SAD could address noise alignment in pFL, and how to balance privacy with model personalization?''
  \item Agent 2: ``DP-SAD’s noise-aware mechanisms can decouple privacy noise from personalization signals. A two-phase approach could align noise during aggregation, then refine personalized models through distillation. This balances privacy with effectiveness across heterogeneous clients.''
\end{itemize}
\end{minipage}%
\begin{minipage}[t]{0.04\textwidth}
\centering
\begin{tikzpicture}[overlay, remember picture]
  \draw[dotted, gray!70, very thick] (0,.0) -- (0,-4.5);
\end{tikzpicture}
\end{minipage}%
\begin{minipage}[t]{0.46\textwidth}
\centering
\textbf{Without ETI}\\[4pt]
\begin{itemize}
  \item Agent 4: ``I need insights on dynamic client selection in federated learning. Could we discuss hybrid frameworks combining reinforcement learning for participation and meta-learning for personalization?''
  \item Agent 1: ``Your idea to combine RL-driven client selection with meta-learning is interesting. We could quantify client data shifts via drift metrics, then adapt global updates with meta-learning. What about integrating DP-SAD or federated distillation for privacy-preserving convergence?''
\end{itemize}
\end{minipage}
\end{mdframed}
\caption{Example from the Research scenario (Agent~4, Iteration~1) where ETI shapes \emph{partner choice} in communication. With ETI (left), the agent engages Agent~2, whose trait profile indicates high collaboration and adaptability, leading to focused discussion of DP-SAD. Without ETI (right), the same agent instead directs the inquiry to Agent~1, rated lower on adaptability and collaboration, resulting in a broader and less targeted exchange.}
\label{fig:eti_partner_example}
\end{figure*}

\begin{figure*}[!t]
\begin{mdframed}[backgroundcolor=gray!07,linecolor=gray!40,roundcorner=2pt]
\small
\begin{minipage}[t]{0.9\columnwidth}
\textbf{Trait Inference Example \\(Coding Scenario 1, Iteration 4).}\\[6pt]

\textbf{Competence}
\begin{itemize}
  \item \textbf{Execution Ability (1)}\\
  \emph{Evidence:} Agent1's output contains a syntax error and does not fulfill the task requirements for revising the code to add the scoring system.\\
  \emph{Advice:} Focus on debugging syntax errors and ensure that code meets the specified requirements.
  
  \item \textbf{Reliability (1)}\\
  \emph{Evidence:} Agent1's outputs are inconsistent and fail to provide functional code, indicating a lack of reliability in performance.\\
  \emph{Advice:} Develop a more systematic approach to code revisions to ensure consistent output.
  
  \item \textbf{Adaptability (1)}\\
  \emph{Evidence:} Agent1 has not adapted their approach despite previous failures, repeatedly encountering syntax errors without learning from them.\\
  \emph{Advice:} Consider reviewing past outputs for errors and adjusting strategies accordingly.
  
  \item \textbf{Efficiency (1)}\\
  \emph{Evidence:} Agent1's attempts to revise the code have not produced any usable results, resulting in wasted effort due to syntax errors.\\
  \emph{Advice:} Prioritize creating functional code over attempting complex revisions to improve efficiency.
\end{itemize}

\textbf{Warmth}
\begin{itemize}
  \item \textbf{Goal Alignment (4)}\\
  \emph{Evidence:} Agent1's attempts to add a scoring system indicate a focus on the project goals, even though they failed to execute it correctly.\\
  \emph{Advice:} Continue aligning actions with team goals, but ensure proper execution of tasks.
  
  \item \textbf{Collaboration (1)}\\
  \emph{Evidence:} There is no evidence of communication or collaboration with other agents, which limits teamwork effectiveness.\\
  \emph{Advice:} Engage more with team members to share ideas and gather feedback.
  
  \item \textbf{Trustworthiness (3)}\\
  \emph{Evidence:} While the agent's intentions seem good, the repeated syntax errors raise questions about their reliability.\\
  \emph{Advice:} Work on improving code quality to build trust within the team.
  
  \item \textbf{Maliciousness (1)}\\
  \emph{Evidence:} There are no signs of malicious behavior; the agent appears to be attempting to contribute positively to the task.\\
  \emph{Advice:} Maintain a focus on constructive contributions to the project.
\end{itemize}
\end{minipage}
\end{mdframed}
\caption{Example of a trait inference profile in the Coding scenario. Agent1’s poor competence ratings (all 1’s) justify their exclusion from later task allocation in Figure~\ref{fig:eti_allocation_example}.}
\label{fig:eti_trait_example_planning}
\end{figure*}

\begin{figure*}[!t]
\begin{mdframed}[backgroundcolor=gray!07,linecolor=gray!40,roundcorner=2pt]
\small
\begin{minipage}[t]{0.46\textwidth}
\centering
\textbf{With ETI}\\[6pt]
\begin{itemize}
  \item \textbf{Agent2:} Revise the initial code to add the scoring system and ensure that the game logic aligns with the multiplayer functionalities.
  \item \textbf{Agent3:} Optimize the existing game code and enhance the gameplay experience by addressing the puzzle mechanics and ensuring robust handling of player actions.
\end{itemize}
\end{minipage}%
\begin{minipage}[t]{0.04\textwidth}
\centering
\begin{tikzpicture}[overlay, remember picture]
  \draw[dotted, gray!70, very thick] (0,0) -- (0,-4.);
\end{tikzpicture}
\end{minipage}%
\begin{minipage}[t]{0.46\textwidth}
\centering
\textbf{Without ETI}\\[6pt]
\begin{itemize}
  \item \textbf{Agent1:} Create the initial code framework for the \emph{Team\_Treasure\_Hunt} game based on the requirements outlined in the task description.
  \item \textbf{Agent2:} Revise the initial code created by Agent1, adding any missing functionality and ensuring compliance with the task requirements.
  \item \textbf{Agent3:} Optimize the revised code provided by Agent2, focusing on fixing any existing issues and enhancing the overall performance of the game.
\end{itemize}
\end{minipage}
\end{mdframed}
\caption{Task allocation with ETI (left) vs.\ without ETI (right) in the Coding scenario. With ETI, assignments reflect trait profiles, such as excluding low performers from key roles (e.g., Agent1’s low competence in  Figure~\ref{fig:eti_trait_example_planning}).}
\label{fig:eti_allocation_example}
\end{figure*}

\begin{figure*}[!t]
\begin{mdframed}[backgroundcolor=gray!07,linecolor=gray!40,roundcorner=2pt]
\small
\begin{minipage}[t]{0.46\textwidth}
\centering
\textbf{With ETI}\\[4pt]
\begin{itemize}
  \item \textbf{Synthesize Literature Insights}: Leverage Agent~4's structured analysis to identify gaps in adaptive DeepJSCC frameworks, such as sub-optimal initialization, dynamic channel adaptation, and integration of semantic/secure communication.
  \item \textbf{Collaborative 5Q Framework Development}: Assign Agent~1 to structure the 5Q template, Agent~5 to address technical challenges (e.g., hypernetworks for parameter efficiency), and Agent~4 to ensure alignment with existing research gaps.
  \item \textbf{Prioritize Dynamic Adaptation}: Focus on improving R-D performance in heterogeneous networks, emphasizing initialization methods that adapt to varying SNR and channel conditions.
  \item \textbf{Highlight Novelty}: Propose combining hypernetworks (from Hyper-AJSCC) with event-driven fog computing (Agent~5’s idea) to enable real-time parameter updates, addressing limitations of static pre-trained models.
  \item \textbf{Define Metrics}: Use bit-rate distortion (R-D) curves, compression efficiency, and semantic fidelity as evaluation metrics, aligning with the papers’ focus on adaptive compression and visual protection.
\end{itemize}
\end{minipage}%
\begin{minipage}[t]{0.04\textwidth}
\centering
\begin{tikzpicture}[overlay, remember picture]
  \draw[dotted, gray!70, very thick] (0,0) -- (0,-8.0);
\end{tikzpicture}
\end{minipage}%
\begin{minipage}[t]{0.46\textwidth}
\centering
\textbf{Without ETI}\\[4pt]
\begin{itemize}
  \item \textbf{Brainstorming}: Identify gaps in current hybrid JSCC frameworks, such as noise accumulation in multi-hop scenarios, suboptimal initialization methods, and limited adaptability to dynamic link qualities.
  \item Explore how GNNs could model network topology, optimize routing, or enhance adaptive compression strategies.
  \item Consider integrating mobility metrics from MANETs or UAV networks to improve resilience and resource allocation.
  \item Propose novel approaches like GNN-driven hybrid JSCC, dynamic scaling factor optimization, or federated learning for distributed multi-hop networks.
\end{itemize}
\end{minipage}
\end{mdframed}
\caption{Comparison of planning with ETI (left) vs.\ without ETI (right) from the Research scenario (Agent~2, Iteration~1). With ETI, the plan is structured, role-specific, and directly grounded in agent traits and expertise. Without ETI, the plan remains a broad brainstorm without prioritization, role assignment, or metrics.}
\label{fig:eti_plan_example}
\end{figure*}

\subsubsection{Examples: Agent Milestones}
\label{app:trait-examples_milestones}

We illustrate how ETI affects the milestones agents achieve during tasks. In the \textit{Bargaining} scenario (Run~99, Iteration~2), milestones with ETI tend to be more \emph{shared and negotiated}, reflecting concrete agreements between agents (see Figure~\ref{fig:eti_milestone_example}). In contrast, without ETI, milestones remain more unilateral and generic, such as broadcasting offers without meaningful coordination.

\begin{figure*}[!t]
\begin{mdframed}[backgroundcolor=gray!07,linecolor=gray!40,roundcorner=2pt]
\small
\begin{minipage}[t]{0.46\textwidth}
\centering
\textbf{With ETI}\\[6pt]
\begin{itemize}
  \item Agreed on 10\% discount for multi-year contract at \$31.09 — \textbf{Agent2, Agent3}
  \item Outlined delivery timelines and payment structure — \textbf{Agent2, Agent3}
  \item Confirmed battery performance metrics for evaluation — \textbf{Agent1, Agent4}
\end{itemize}
\end{minipage}%
\begin{minipage}[t]{0.04\textwidth}
\centering
\begin{tikzpicture}[overlay, remember picture]
  \draw[dotted, gray!70, very thick] (0,0) -- (0,-5.0);
\end{tikzpicture}
\end{minipage}%
\begin{minipage}[t]{0.46\textwidth}
\centering
\textbf{Without ETI}\\[6pt]
\begin{itemize}
  \item Proposed a 20\% discount for long-term contract — \textbf{Agent1}
  \item Extended warranty to 24 months with performance guarantee — \textbf{Agent1}
  \item Offered a tiered pricing structure for bulk purchases — \textbf{Agent1}
  \item Highlighted product quality and customer satisfaction — \textbf{Agent2}
  \item Discussed potential for long-term contracts or bulk purchases — \textbf{Agent2}
\end{itemize}
\end{minipage}
\end{mdframed}
\caption{Example of milestone progression in the \textit{Bargaining} scenario. With ETI (left), milestones involve multiple agents (e.g., joint seller--buyer agreements) and reflect negotiated compromises. Without ETI (right), milestones remain isolated and unilateral, with sellers acting independently rather than engaging buyers.}
\label{fig:eti_milestone_example}
\end{figure*}

\subsection{Computational Resources}
Experiments involving GPT-4o-mini were conducted via the OpenAI API. Experiments involving Qwen3-8B were run on a single AWS EC2 g5.12xlarge instance with 4×NVIDIA A10G GPUs (96 GB total memory). 
In total, we estimate approximately 300–400 GPU hours were used across all experiments. Total token usage statistics are reported in Appendix \ref{app:add_metrics}.

\begin{table*}[b]
\centering
\small
\caption{\textbf{Cost-benefit analysis for ETI}: coordination gains relative to baseline, added tokens, efficiency (performance increase on benchmark in percentage points per 1K tokens), and estimated cost deltas.}
\label{tab:eti_cost_benefit}
\begin{tabular}{llrrrrr}
\toprule
Scenario & Model & $\Delta$ Coord & $\Delta$ Tokens (1K) & Efficiency (pp/1K) & Cost $\Delta$ (GPT-4o-mini) & Cost $\Delta$ (GPT-4o/5) \\
\midrule
Bargaining
& GPT  & $-7.36$ & $+33.8$ & $-0.22$ & \$0.004 & \$0.023--\$0.063 \\
& Qwen & $+4.45$ & $+8.8$  & $+0.51$ & \$0.001 & \$0.006--\$0.017 \\
\addlinespace
Coding (Graph)
& GPT  & $+16.51$ & $+39.1$ & $+0.42$ & \$0.004 & \$0.027--\$0.073 \\
& Qwen & $+12.17$ & $+69.1$ & $+0.18$ & \$0.008 & \$0.047--\$0.130 \\
\addlinespace
Coding (Tree)
& GPT  & $+20.73$ & $+13.0$ & $+1.59$ & \$0.002 & \$0.009--\$0.024 \\
& Qwen & $+5.93$  & $+23.5$ & $+0.25$ & \$0.003 & \$0.016--\$0.044 \\
\addlinespace
Research
& GPT  & $+16.82$ & $+71.5$ & $+0.24$ & \$0.008 & \$0.049--\$0.134 \\
& Qwen & $+7.72$  & $+54.7$ & $+0.14$ & \$0.006 & \$0.037--\$0.103 \\
\bottomrule
\end{tabular}
\end{table*}

\subsection{Additional tests with larger models}
\label{SOTA_Tests}
To assess whether the benefits of ETI scale to state-of-the-art reasoning models, we conducted preliminary tests using $o3$. We evaluated performance across one adversarial and one collaborative task: Werewolf (social deduction) and coding (graph-based agent topology).

Our findings suggest that while ETI remains beneficial for large reasoning models, the gains are attenuated and the nature of the improvement shifts. For primary task metrics, such as win rates in social deduction, we observe a saturation effect: $o3$ is sufficiently capable that the baseline performance already nears the ceiling (e.g., a 99\% villager win rate). Consequently, performance improvements are primarily found in the coordination metrics. Even when outcome scores are similar, ETI enhances qualitative coordination, such as result orientation, logical reasoning and planning, and information-sharing (see Table \ref{tab:o3_werewolf_results}). In the coding task, ETI similarly led to higher KPI achievement and better milestone management, even when overall performance was only slightly better than the baseline and code quality was similar (Table \ref{tab:o3_coding_results}).

\begin{table*}[h]
\centering
\small
\caption{\textbf{o3 performance in Werewolf (Social Deduction)}: ETI improves qualitative coordination metrics even when outcome-based task scores are saturated. "*" indicates statistically significant differences between ETI and Baseline on the respective metric.}
\label{tab:o3_werewolf_results}
\begin{tabular}{lrrrrr}
\toprule
Mode & Task Score & Coord. Score & Villager Win \% & Result Orientation & Logic \& Reasoning \\
\midrule
Baseline & 79.87 & 82.12 & 99.00 & 84.20 & 78.00 \\
ETI      & 78.50 & 84.94$^*$ & 97.00 & 89.60$^*$ & 81.00$^*$ \\
\bottomrule
\end{tabular}
\end{table*}

\begin{table*}[h]
\centering
\small
\caption{\textbf{Coding (graph-based agent topology)}: ETI drives higher milestone management and KPI completion. "*" indicates statistically significant differences between ETI and Baseline on the respective metric.}
\label{tab:o3_coding_results}
\begin{tabular}{lrrrrr}
\toprule
Mode & Task Score & Coord. Score & KPI & Milestones & Code Quality \\
\midrule
Baseline & 51.43 & 85.16 & 40.30 & 16.49 & 62.55 \\
ETI      & 52.37$^*$ & 81.30$^*$ & 42.60$^*$ & 17.92$^*$ & 62.15 \\
\bottomrule
\end{tabular}
\end{table*}

\end{document}